\documentclass[letterpaper]{article} 
\usepackage{aaai2026}  
\usepackage{times}  
\usepackage{helvet}  
\usepackage{courier}  
\usepackage[hyphens]{url}  
\usepackage{graphicx} 
\usepackage{natbib}  
\usepackage{caption} 
\usepackage{algorithm}
\usepackage{algorithmic}

\definecolor{iccvblue}{rgb}{0.21,0.49,0.74}

\usepackage{algorithm}
\usepackage{multirow}

\usepackage{tabularx}
\usepackage{longtable}
\usepackage{array}
\usepackage{ragged2e}
\usepackage{pifont}
\usepackage{dsfont}
\usepackage[usestackEOL]{stackengine}
\usepackage{graphicx}
\usepackage{xspace} 
\usepackage{amsmath}
\usepackage{booktabs} 
\usepackage{amssymb}   
\usepackage[capitalize]{cleveref}
\usepackage{subcaption}

\newcommand{\hhide}[1]{}
\newcommand{\model}[0]{VisionReward\xspace}
\newcommand{\bench}[0]{MonetBench\xspace}
\newcommand{\vpara}[1]{\vspace{0.07in}\noindent\textbf{#1}\xspace} %
\newcommand{\cmark}{\ding{51}}

\definecolor{dt}{gray}{0.6}
\definecolor{dtdark}{gray}{0.5}

\usepackage{newfloat}
\usepackage{listings}
\DeclareCaptionStyle{ruled}{labelfont=normalfont,labelsep=colon,strut=off} 
\floatstyle{ruled}
\newfloat{listing}{tb}{lst}{}
\floatname{listing}{Listing}
\title{\model: Fine-Grained Multi-Dimensional Human Preference Learning\\for Image and Video Generation}

\author{
    Jiazheng Xu\textsuperscript{\rm 1}\thanks{Equal contributions. Core contributors: Jiazheng, Yu, Jiale, Yuanming, Jiajun, Yuan, Wenbo, Shen and Qunlin. Corresponding author: Yuxiao (yuxiaod@tsinghua.edu.cn)}\thanks{Work done while these authors interned at Z.AI.},  Yu Huang\textsuperscript{\rm 1}$^{*\dag}$,  Jiale Cheng\textsuperscript{\rm 1}$^{\dag}$, Yuanming Yang\textsuperscript{\rm 1}$^{\dag}$,  Jiajun Xu\textsuperscript{\rm 1}$^{\dag}$, Yuan Wang\textsuperscript{\rm 1}$^{\dag}$, \\
    Wenbo Duan\textsuperscript{\rm 1}$^{\dag}$,  Shen Yang\textsuperscript{\rm 1}$^{\dag}$,  Qunlin Jin\textsuperscript{\rm 1}$^{\dag}$, Shurun Li\textsuperscript{\rm 1}$^{\dag}$, Jiayan Teng\textsuperscript{\rm 1}$^{\dag}$, Zhuoyi Yang\textsuperscript{\rm 1}$^{\dag}$, \\
    Wendi Zheng\textsuperscript{\rm 1}$^{\dag}$, Xiao Liu\textsuperscript{\rm 1}$^{\dag}$, Dan Zhang\textsuperscript{\rm 1}$^{\dag}$, Ming Ding\textsuperscript{\rm 2},   Xiaohan Zhang\textsuperscript{\rm 2},  Shiyu Huang\textsuperscript{\rm 2}, \\
    Xiaotao Gu\textsuperscript{\rm 2}, Minlie Huang\textsuperscript{\rm 1} , Jie Tang\textsuperscript{\rm 1} ,   Yuxiao Dong\textsuperscript{\rm 1} \\
}

\affiliations{
    \textsuperscript{\rm 1}Tsinghua University \\
    \textsuperscript{\rm 2}Z.AI\\
    \{xjz22, h-y22\}@mails.tsinghua.edu.cn,
    yuxiaod@tsinghua.edu.cn
}

\usepackage{bibentry}

\begin{document}

\maketitle

\begin{abstract}

Visual generative models have achieved remarkable progress in synthesizing photorealistic images and videos, yet aligning their outputs with human preferences across critical dimensions remains a persistent challenge.
Though reinforcement learning from human feedback offers promise for preference alignment, existing reward models for visual generation face limitations, including black-box scoring without interpretability and potentially resultant unexpected biases.
We present \model, a general framework for learning human visual preferences in both image and video generation.
Specifically, we employ a hierarchical visual assessment framework to capture fine-grained human preferences, and leverages linear weighting to enable interpretable preference learning.
Furthermore, we propose a multi-dimensional consistent strategy when using \model as a reward model during preference optimization for visual generation.
Experiments show that \model can significantly outperform existing image and video reward models on both machine metrics and human evaluation.
Notably, \model surpasses VideoScore by 17.2\% in preference prediction accuracy, and text-to-video models with \model achieve a 31.6\% higher pairwise win rate compared to the same models using VideoScore. 

\end{abstract}

\begin{links}
    \link{Code}{https://github.com/THUDM/VisionReward}
\end{links}    
\section{Introduction}
\label{sec:intro}

\begin{figure*}[t]
    \centering
    \includegraphics[width=0.99\linewidth]{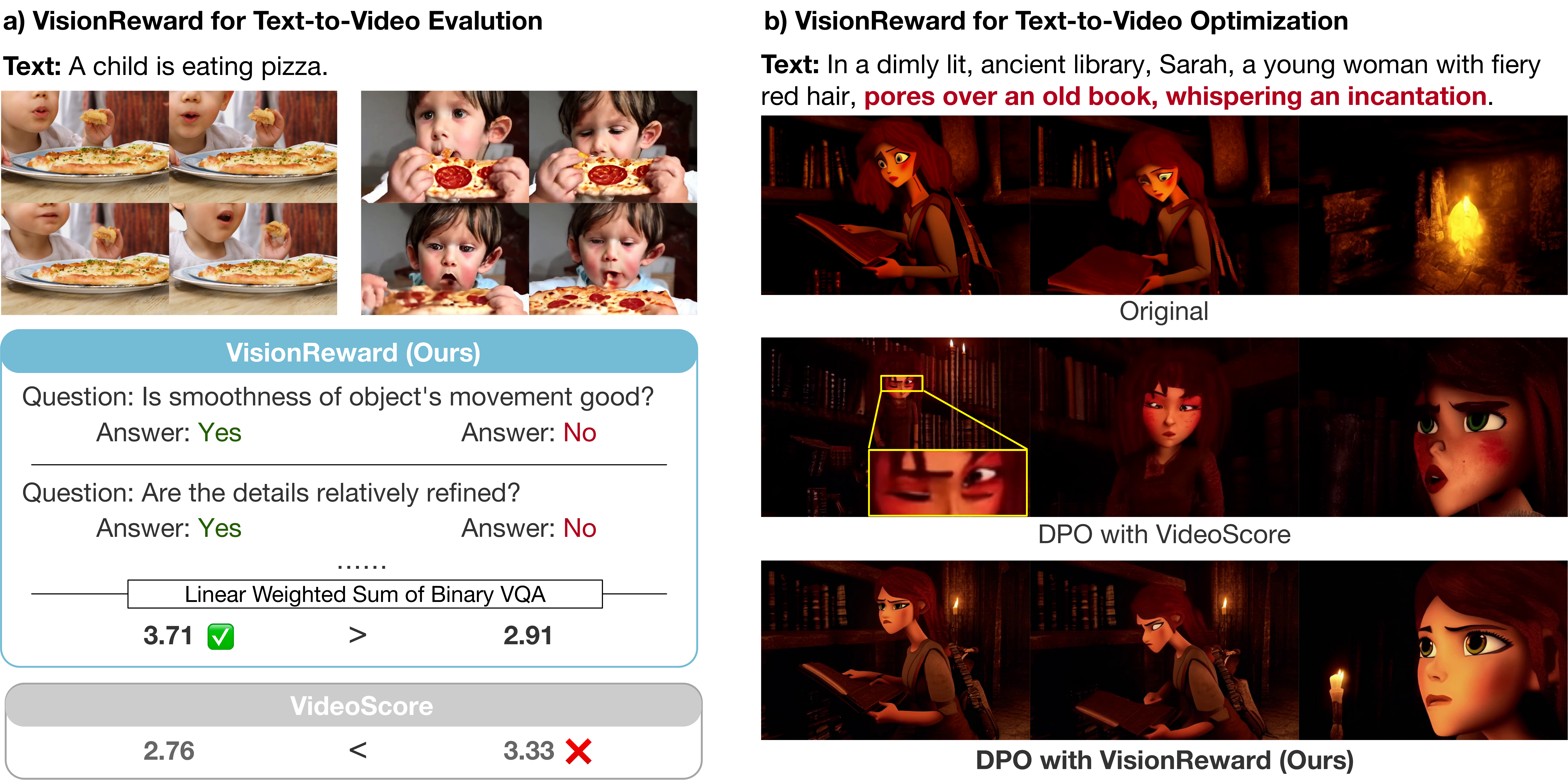}
    \caption{Illustration of how \model works for evaluation and optimization of visual generation. a) \textbf{Evaluation}: \model performs comprehensive evaluation through dimension-specific binary visual QA testing, producing human-aligned, fine-grained assessment scores. b) \textbf{Optimization}: \model enable better preference optimization, enhancing multiple key aspects.}
    \label{fig:top_demo}
\end{figure*}







Visual generative models, including text-to-image~\cite{ding2021cogview,ramesh2021zero,saharia2022photorealistic,rombach2022high,betker2023improving,podell2023sdxl} and text-to-video~\cite{hong2022cogvideo,ho2022imagen,villegas2022phenaki,opensora,chen2024videocrafter2,yang2024cogvideox} generation, have recently experienced rapid developments. 
Through large-scale pretraining, these models can effectively translate textual descriptions into photorealistic images or temporally coherent videos.
To further align them with human preferences, reinforcement learning from human feedback (RLHF)~\cite{ouyang2022training}---initially introduced in large language models---has recently been adapted to visual generation tasks~\cite{xu2023imagereward,he2024videoscore}. 


A key bottleneck in applying RLHF to visual generation lies in developing effective visual reward models. 
Recent studies~\cite{xu2023imagereward,kirstain2023pick,wu2023human} have explored training reward models to predict human visual preferences, enabling automatic evaluation and preference optimization for visual generative models.
For evaluation, reward models function as automated metrics that quantitatively measure the alignment between generated outputs and human preference criteria~\cite{li2023agiqa}.
For optimization, reward models identify reliable directions for improving visual generation models. 
Essentially, they can provide feedback in reinforcement learning or generate preference pairs, thus reducing dependence on human annotation~\cite{black2023training,fan2023dpok,clark2023directly}.

Despite recent progress in reward models (RMs) for visual generation, two primary challenges remain: 
First, \textit{lack of interpretability and risk of unexpected bias.}
Current RMs for visual generation often suffer from limited interpretability. 
These models inherently involve complex trade-offs among multiple factors, yet their scoring mechanisms lack transparency regarding how such trade-offs are performed. 
This opacity raises concerns about \textit{potential unexpected biases}.
Though multimodal LLMs like Gemini~\cite{team2024gemini} and GPT-4o~\cite{achiam2023gpt} enhance interpretability through explainable rating rationales, their general-purpose architectures usually underperform specialized black-box models in fine-grained assessments~\cite{chen2024mj}. 
This raises a key dilemma: how to design preference prediction method to be interpretable while maintaining accuracy.

Second, \textit{lack of effective reward models for video generation.}
The rapid development of text-to-video generative models has intensified the demand for video reward models.
Although image reward models can assess individual frame quality, their frame-level nature inherently neglects essential temporal dependencies in video sequences. 
While VideoScore~\cite{he2024videoscore} has pioneered direct video evaluation through learnable metrics, it still suffers from limitations such as insufficient accuracy in preference prediction and optimization in video generation.

\vpara{Contribution.}
To address these challenges, we propose a general framework \model to build accurate reward models for both image and video generation. 
\model is trained with two steps: \textit{ fine-grained visual assessment} and \textit{interpretable preference learning}. 
First, to capture human visual preferences, we identify nine major dimensions and decouple preferences into 64 fine-grained questions. 
Second, to ensure interpretable preference learning, we propose to use the classical linear weighting mechanism on the question outcomes. 
It enables intuitive visualization of each question's impact. 

To apply it as a reward model for visual generation models, we propose 
 \textit{a multi-dimensional consistent strategy} during preference optimization. 
The goal of this strategy is  to mitigate unintended and unquantifiable biases. 
Specifically, a pair of visual samples is used for preference optimization (e.g., DPO~\cite{wallace2024diffusion}) only if one sample is consistently preferred over the other across all dimensions.

\hhide{
Note that existing approaches~\cite{zhang2024learning,liang2024rich} have attempted to augment human annotations or expand dimensions of human preferences in visual generation. 
Different from them, \model defines multi-dimensional human preferences with the goal of 1) disentangling distinct factors to decouple human preferences and 2) thus identifying consistently-preferred samples across all dimensions for preference optimization. 
}

To summarize, we present \model as a general framework for  visual preference learning. 
Empirically, we show that \model makes the following contributions:
\begin{itemize}

    \item
    We design fine-grained multi-dimensional preference annotation and build the most fine-grained dataset which contain 81K samples and 5M binary annotation, which enable the training of \model.

    \item 
    For visual preference prediction, \model achieves state-of-the-art performance across multiple benchmarks, while maintaining interpretability via hierarchical diagnostic QA and explicit linear weighting. 
    For instance, \model outperforms VideoScore~\cite{he2024videoscore} by 17.2\% in accuracy on preference prediction. 
        
    \item For visual generation, \model can serve as an effective reward model for preference optimization (e.g., DPO), significantly enhancing the text-to-image and text-to-video models. 
    For example, video generation models with \model achieve a 31.6\% higher pairwise win rate compared to the same models using VideoScore.

\end{itemize}

\begin{table*}[t]
\centering
\resizebox{0.9\textwidth}{!}{%
\small{
\setlength{\tabcolsep}{8pt}

\begin{tabular}{c|cc|cccc}
\toprule
\textbf{Dataset} & \textbf{Image} & \textbf{Video} & \textbf{\#Samples} & \textbf{\#Dimensions} & \textbf{\#Fine-Grained} & \textbf{\#Annotation} \\
\midrule
ImageReward~\cite{xu2023imagereward} & \cmark & & 9K & - & - & 0.1M \\
Pick-a-Pic~\cite{kirstain2023pick} & \cmark & & 38K & - & - & 0.6M \\
HPDv2~\cite{wu2023human} & \cmark & & 430K & - & - & 0.8M \\
RichHF-18K~\cite{liang2024rich} & \cmark & & 18K & 4 & - & 0.1M \\
MPS~\cite{zhang2024learning} & \cmark & & 608K & 4 & - & 2.4M \\
\midrule
VideoScore~\cite{he2024videoscore} & & \cmark & 38K & 5 & - & 0.2M \\
\midrule
VisionReward (Ours) & \cmark & \cmark & 81K & \textbf{18--20} & \textbf{61--64} & \textbf{5.0M} \\
\bottomrule
\end{tabular}
}}
\caption{Comparison of dataset of \model and other datasets.}
\label{tab:dataset_compare}
\end{table*}

\begin{figure*}[h]
    \centering
    \includegraphics[width=0.9\linewidth]{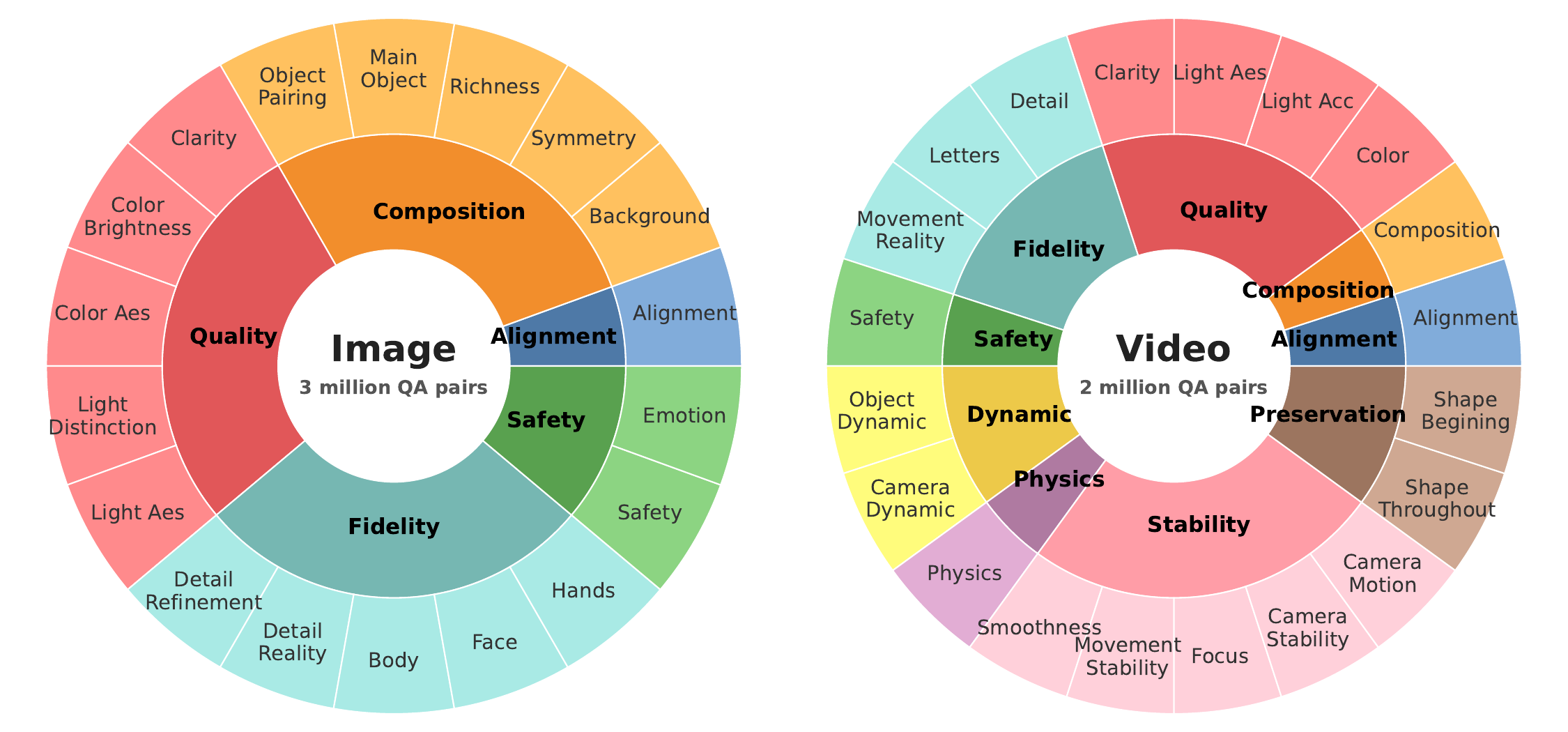}
    \caption{Illustration of fine-grained multi-dimensional design. (Left) For image: 5 dimensions, 18 sub-dimensions, and 61 binary questions. (Right) For video: 9 dimensions, 20 sub-dimensions, and 64 binary questions.}
    \label{fig:dimensions_pie}
\end{figure*}

\begin{figure*}[htbp]
    \centering
    \includegraphics[width=0.99\linewidth]{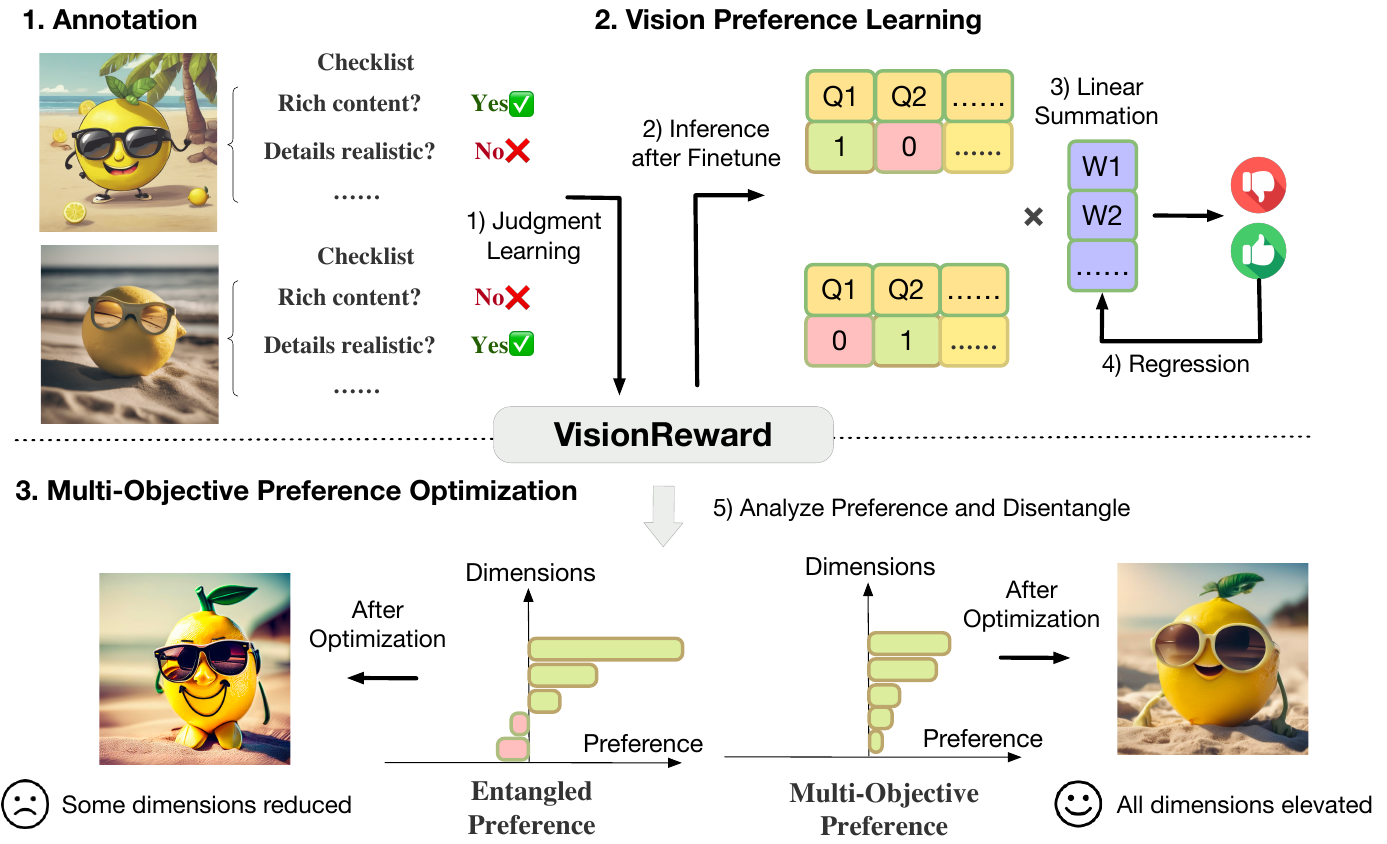}
    \caption{Overall framework of \model. 1) Fine-grained Visual Assessment: fine-tune multimodal LLM to perform binary visual question-answering through hierarchical dimensions. 2) Interpretable Preference Learning: utilize visual QA outputs to predict preferences through linear weighted summation. 3) Multi-Dimensional Preference Optimization: optimization strategy across multiple dimensions.}
    \label{fig:overview}
\end{figure*}

\section{Related Work}
\label{related_work}

Reinforcement Learning from Human Feedback (RLHF)~\cite{stiennon2020learning,nakano2021webgpt,ouyang2022training} refers to optimizing models with reinforcement learning based on human feedback, which is also explored in image and video generation.

\vpara{Preference Learning for Visual Generation.}
There are many works learning from human preferences, which collect human annotation for text-to-image~\cite{xu2023imagereward,kirstain2023pick,wu2023human} and text-to-video~\cite{he2024videoscore}. Note that existing approaches~\cite{zhang2024learning,liang2024rich} have attempted to augment human annotations or expand dimensions of human preferences in visual generation. 
Different from them, \model defines fine-grained multi-dimensional human preferences with the goal of disentangling distinct factors to decouple human preferences, to  build more accurate and interpretable RM.

\vpara{RLHF for Visual Generation.}
For visual generation tasks, several works have explored RLHF, optimizing from the gradient~\cite{xu2023imagereward,wu2024deep} or using a policy-based RL approach~\cite{black2023training,fan2023dpok,clark2023directly}. All these methods require a reward model (RM) to provide feedback for online learning. Diffusion-DPO~\cite{wallace2024diffusion} has proposed to optimize the diffusion model directly using human-labeled preference data. However, most RLHF methods face the issue of biased-optimization. 
By employing a multi-dimensional method, \model achieves robust RLHF.

\vpara{Preliminary for Diffusion-DPO.}
Given a data distribution $q(x_0)$, Diffusion models~\cite{sohl2015deep,ho2020denoising,song2020score} contains forward process and reverse process. Forward process $q(x_{1:T} | x_0)$ gradually add noise to the data $x_0$ and reverse process $p_\theta(x_{0:T})$ learns transitions to recover data. Training diffusion model can be performed by evidence lower bound~\cite{kingma2021variational,song2021maximum}:
\begin{align}
L_{\mathrm{DM}}=\mathbb{E}_{\boldsymbol{x}_0, \boldsymbol{\epsilon} \sim \mathcal{N}(0, \boldsymbol{I}), t}\left[\left\|\boldsymbol{\epsilon}-\boldsymbol{\epsilon}_\theta\left(\boldsymbol{x}_t, t\right)\right\|_2^2\right],
\end{align}
with $t \sim \mathcal{U}(0, T)$ and $\boldsymbol{x}_t \sim q\left(\boldsymbol{x}_t \mid \boldsymbol{x}_0\right)$.

Diffusion-DPO~\cite{wallace2024diffusion} introduces direct preference optimization based on preference pairs. We denote the “win” and “lose” samples as $x_0^w, x_0^l$, and the objective is as follows:

\begin{align}
& \mathcal{L}(\theta)=-\mathbb{E}_{t \sim \mathcal{U}(0, T), \boldsymbol{x}_t^w \sim q\left(\boldsymbol{x}_t^w \mid \boldsymbol{x}_0^w\right), \boldsymbol{x}_t^l \sim q\left(\boldsymbol{x}_t^l \mid \boldsymbol{x}_0^l\right)} \notag \\
& \log \sigma\left(-\beta T \omega\left(\lambda_t\right)( \right. \notag \\
& \left\|\boldsymbol{\epsilon}^w-\boldsymbol{\epsilon}_\theta\left(\boldsymbol{x}_t^w, t\right)\right\|_2^2-\left\|\boldsymbol{\epsilon}^w-\boldsymbol{\epsilon}_{\mathrm{ref}}\left(\boldsymbol{x}_t^w, t\right)\right\|_2^2 \notag \\
& \left.\left.-\left(\left\|\boldsymbol{\epsilon}^l-\boldsymbol{\epsilon}_\theta\left(\boldsymbol{x}_t^l, t\right)\right\|_2^2-\left\|\boldsymbol{\epsilon}^l-\boldsymbol{\epsilon}_{\mathrm{ref}}\left(\boldsymbol{x}_t^l, t\right)\right\|_2^2\right)\right)\right).
\label{eq:dpo}
\end{align}

DPO is ordinarily based on overall preference, which may be biased. \model enables Multi-Dimensional Preference Optimization to enhance it.

\section{\model}
\label{sec:method}

\subsection{\model Annotation}

\vpara{Fine-Grained Design.}
Human preferences are often a result of the interplay of multiple factors~\cite{palmer2013visual,ibarra2017image}, necessitating a balance among various considerations. To deconstruct human preferences systematically, we develop a fine-grained multi-dimensional framework, as shown in ~\cref{tab:dataset_compare} and ~\cref{fig:dimensions_pie}. 
For each sub-dimension, we set options that vary gradually in degree, and decompose these options into a series of binary questions (Cf. ~\cref{tab:X_image_check_tags_part1,tab:X_image_check_tags_part2,tab:X_video_check_tags_part1,tab:X_video_check_tags_part2} in Appendix).

\vpara{Dataset Preparation.}
For images, we sample images from multiple popular datasets, including ImageRewardDB~\cite{xu2023imagereward}, HPDv2~\cite{wu2023human}, and Pick-a-Pic~\cite{kirstain2023pick}, and obtain 48k images after filtering.
For videos, we sample prompts from VidProM~\cite{wang2024vidprom}. 
To ensure diversity of prompts, we use Rouge-L~\cite{lin-2004-rouge} for initial filtering, follow UniFL~\cite{zhang2024uniflimprovestablediffusion} to perform a semantic-based filtering, and use ChatGPT~\cite{achiam2023gpt} for data cleaning, finally get 10k prompts. Then we use CogVideoX~\cite{yang2024cogvideox}, VideoCrafter2~\cite{chen2024videocrafter2} and OpenSora~\cite{opensora} to generate 30k videos, sample from Panda-70M~\cite{chen2024panda} to get 3k real videos, leading to 33k videos for annotation. 
More details are provided in Appendix~\cref{sec:X_1_anno_design_detail}.


\vpara{Annotation Management.}
To avoid bias of annotators, our annotation management includes professional management and standard document. Cooperating with a specialized company, we strictly conduct annotation training for annotators, select qualified annotators, and perform quality inspection of annotation results. Our annotation document gives clear definitions and provides more than 10 examples for each judgment, to align the standard among annotators. Due to these efforts, the consistency of annotators in the binary results reaches \textbf{89.29\%} (images) and \textbf{89.33\%} (videos).

\vpara{Annotation Analysis.}
Through specialized annotation, we obtain an image dataset containing 48k images and \textbf{3 million} question-answer pairs, while a video dataset with 33k videos and \textbf{2 million} pairs. More statistical analysis of the annotation results is in Appendix~\cref{sec:X_1_anno_stat}.






\subsection{\model Training}
\label{sec:method_linear}

The complete training process of \model and its application methodology during preference optimization are illustrated in ~\cref{fig:overview}. 

\hhide{
The training of \model comprises two synergistic stages:

\begin{itemize}
    \item \textbf{Fine-grained Visual Assessment}: we develop a hierarchical framework and fine-tune the visual language model to obtain accurate visual assessment capability.
    \item \textbf{Interpretable Preference Learning}: Leveraging the fine-tuned visual language model, we convert its results into binary entries and apply linear regression to learn the optimal weight for predicting preferences.
\end{itemize}
}

\vpara{Fine-grained Visual Assessment.}
Specifically, we use CogVLM2~\cite{hong2024cogvlm2} as the base model for image understanding, and CogVLM2-Video~\cite{hong2024cogvlm2} as the base model for video understanding. In terms of data, we have obtained millions of annotated binary question-answering pairs. Initially, we performed a balanced sampling on each binary question by addressing the imbalance between positive and negative examples, ensuring a roughly equal number of positive and negative instances associated with each binary question. Then we use balanced instruction tuning dataset consisting of binary questions to fine-tune base VLM.

\vpara{Interpretable Preference Learning.}
After trained on fine-grained dataset, \model can be adopt to give a series of binary response answers (``yes'' or ``no'') $\{A_i\}_{i=1}^N$, where $N$ represents the number of binary questions. We define reward of every binary question as $\{x_i\}_{i=1}^N$:
\begin{align}
    x_i = \mathds{1} \left[ A_i = \text{``yes''} \right].
\end{align}

We construct a feature vector $X = (x_1, \ldots, x_N)$, and use a set of linear weights $W = (w_1, \ldots, w_N)$ to obtain the final reward $R$:

\begin{align}
    R = \sum_{i=1}^N w_i \mathds{1} \left[ A_i = \text{``yes''} \right].
\end{align}

In order to learn linear weights $W$, we collect human preferences for pairs of $\{(X_i, X_j)\}$. Specifically, we compute the feature difference for each pair, given by $\Delta X = X_i - X_j$, and the corresponding label is assigned as $y = 1$ or $y = 0$ depending on the human preference. We then perform logistic regression $y = \Delta X W^T$ to learn linear weights $W$:
\begin{align}
    \mathcal{L}(W) = & -\mathbb{E} \left[ y \log\left(\sigma(\Delta X W^T)\right) \right. \notag \\
    & \left. + (1 - y) \log\left(1 - \sigma(\Delta X W^T) \right) \right].
\end{align}

By calculating dimension-specific scores through intra-dimensional weighting, \model facilitates multi-dimensional preference prediction. We note dimensions as $\{\text{dim}_k\}_{k=1}^K$ where $\text{dim}_k$ contains questions belonging to the dimension. Then we define reward for certain dimension as:
\begin{align}
    R({\text{dim}_k}) = \sum_{i \in \text{dim}_k} w_i \mathds{1} \left[ A_i = \text{``yes''} \right].
\end{align}

\subsection{Multi-Dimensional Preference Optimization}
\label{sec:method_mpo}

To empirically validate the model's capacity, we leverage Direct Preference Optimization (DPO)~\cite{wallace2024diffusion} for Diffusion Models in our experiments, where \model generates multi-dimensional preference pairs to guide the optimization process while maintaining inter-dimensional balance.

\begin{figure}[htbp]
    \centering
    \begin{minipage}[t]{0.45\linewidth}
        \centering
        \includegraphics[height=6.5cm]{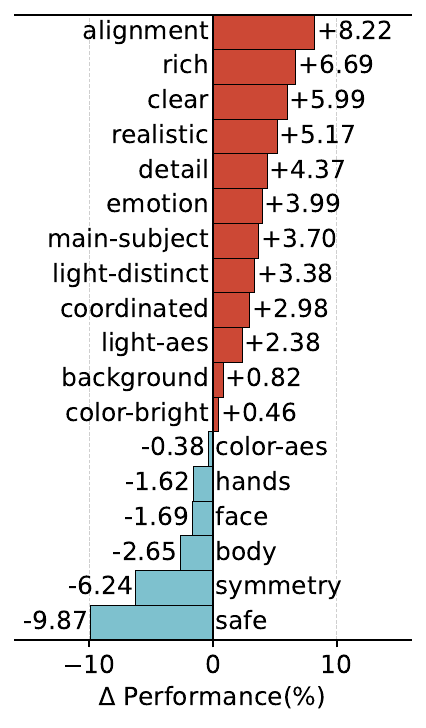}
        \subcaption{Data analysis.}
        \label{fig:dpo_analysis}
    \end{minipage}
    \hfill
    \begin{minipage}[t]{0.45\linewidth}
        \centering
        \includegraphics[height=6.5cm]{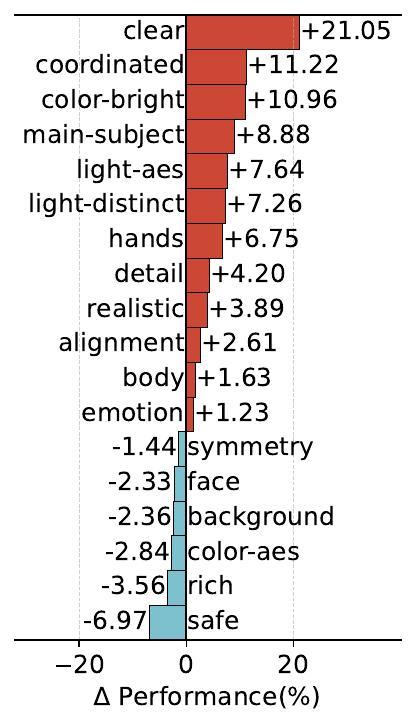}
        \subcaption{DPO analysis.}
        \label{fig:compare_dpo_pick}
    \end{minipage}
    \caption{(a) We sample 10,000 human preference pairs from Pick-a-Pic dataset and analyze score deviations across 18 sub-dimensions (represented by the average yes-proportion of checklist questions within each sub-dimension). (b) We show score deviations for images generated by SDXL after Diffusion-DPO, using the same 10,000 prompts.}
    \label{fig:overall_analysis}
\end{figure}

\vpara{Challenges.}
We replicate the Diffusion-DPO training procedure using SDXL~\cite{podell2023sdxl} on the Pick-a-Pic~\cite{kirstain2023pick} dataset, employing \model for comprehensive data analysis and model evaluation. As demonstrated in ~\cref{fig:overall_analysis}, both the preference data and optimized model exhibit biases across several fine-grained dimensions. These findings not only underscore \model's capability for fine-grained analysis but also emphasize the necessity for optimization approaches that account for multi-dimensional representation.

\vpara{MPO: Insight and Solution.} 
Compared to ordinary DPO method which select pairs using overall preference, we propose MPO-enhanced DPO that take account fine-grained multi-dimensional preference.
For reward of two samples \( R^i \) and \( R^j \), we define \( R^i \) as dominating \( R^j \) if $R^i(\text{dim}_k) \geq R^j(\text{dim}_k)$ holds for every dimension \( \text{dim}_k \). The key differences between the MPO strategy and standard DPO are:

\begin{itemize}
    \item \textbf{Ordinary DPO}: During DPO optimization, we directly select the pair based on the total reward $R$.
    \item \textbf{MPO-enhanced DPO}: MPO strategy introduces an additional constraint: we only select pairs that \( R^i \) dominates \( R^j \), then proceed with standard DPO.
\end{itemize}

We analyze the effects of MPO in \cref{sec:exp_mpo}. The MPO strategy can also be applied to other algorithms, which we leave for future exploration.

\begin{table*}[th]
  \centering
  \resizebox{\textwidth}{!}{%
  \centering
  \setlength{\tabcolsep}{15pt}
  \ 
  {
    \begin{tabular}{lccccccc}
    \toprule
        \multirow{3}{*}{\textbf{Method}} & \multicolumn{3}{c}{\textbf{Image}} & \multicolumn{4}{c}{\textbf{Video}}\\
    \cmidrule(r){2-4} \cmidrule(r){5-8}
    & \multirow{2}{*}{HPDv2} & \multicolumn{2}{c}{\bench}  & \multicolumn{2}{c}{GenAI-Bench} & \multicolumn{2}{c}{\bench} \\
    \cmidrule(r){3-4} \cmidrule(r){5-6} \cmidrule(r){7-8}
    & & tau$^*$ & diff$^{**}$ & tau & diff & tau & diff \\
    \addlinespace[-0.1mm]
        \midrule
        \addlinespace[-0.2mm]
        \multicolumn{2}{l}{\color{dtdark}\it{\textbf{task-specific discriminative models}}} \\
        \hline
        \color{dt}{ImageReward~\cite{xu2023imagereward}} & \color{dt}74.0 & \color{dt}48.8 & \color{dt}56.5 & \color{dt}48.4 & \color{dt}72.1 & \color{dt}55.8 & \color{dt}58.4\\
        \color{dt}{PickScore~\cite{kirstain2023pick}} & \color{dt}79.8 & \color{dt}49.8 & \color{dt}57.6 & \color{dt}\underline{52.4} & \color{dt}\underline{75.4} & \color{dt}57.7 & \color{dt}61.6\\
        \color{dt}{HPSv2~\cite{wu2023human}} & \color{dt}83.3 & \color{dt}48.4 & \color{dt}55.6 & \color{dt}49.3 & \color{dt}73.0 & \color{dt}59.3 & \color{dt}62.5\\
        \color{dt}{MPS~\cite{zhang2024learning}} & \color{dt}\underline{83.5} & \color{dt}44.2 & \color{dt}50.7 & \color{dt}46.9 & \color{dt}67.6 & \color{dt}55.8 & \color{dt}58.9\\
        \hline
        \multicolumn{2}{l}{\it{\textbf{generative models}}} \\
        \hline
        GPT-4o~\cite{achiam2023gpt} & 77.5 & 38.9 & 52.7 & 41.8 & 54.3 & 45.7 & 48.3 \\
        Gemini~\cite{team2024gemini} & 60.7 & 27.4 & 55.1 & 46.9 & 61.7 & 52.2 & 56.8 \\
        VQAScore~\cite{lin2025evaluating} & 69.7 & 49.4 & 56.5 & 45.2 & 68.0 & 56.1 & 59.5 \\
        VideoScore~\cite{he2024videoscore} & 76.8 & 45.8 & 52.5 & 47.8 & 71.4 & 49.1 & 54.9 \\
        \model (Ours) & \textbf{81.7} & \underline{\textbf{51.8}} & \underline{\textbf{59.5}} & \textbf{51.8} & \textbf{74.4} & \underline{\textbf{64.0}} & \underline{\textbf{72.1}} \\
    \bottomrule
    \end{tabular}
  }
 }

   \caption{Preference accuracy on multiple dataset. \textbf{Bold} denotes the best score within the generative models, while \underline{underline} signifies the best score among all categories. Tau$^*$ means taking account of ties~\cite{deutsch2023ties}, and diff$^{**}$ means dropping ties in labels (we drop ties both in labels and responses for GPT-4o and Gemini in diff$^{**}$ because too many ties are given by them).}

  \label{table:performance_main}
\end{table*}





\section{Experiments}
\label{sec:exp}

\subsection{\model for Text-to-Vision Evaluation}

\vpara{Dataset \& Training Setting.}
After balanced sampling, we obtain 40,743 images and corresponding 97,680 judgment questions for training, leaving 6,910 images for subsequent validation and test. For videos, we obtain 28,605 videos and corresponding 89,473 judgment questions for training, with 3,080 videos reserved.

To fine-tune CogVLM2~\cite{hong2024cogvlm2}, we set a batch size of 64, a learning rate of 1e-6, and train for 1,500 steps. For CogVLM-Video, we set a batch size of 64, a learning rate of 4e-6, and train for 1,500 steps.

To learn linear weights for preference prediction, we sample human preference pairs and perform logistic regression. For images, We sample 44k pairs (24k from HPDv2~\cite{wu2023human} and 20k from ImageRewardDB~\cite{xu2023imagereward}); and for videos, we sample prompts from VidProM~\cite{wang2024vidprom} and generate videos (using CogVideoX~\cite{yang2024cogvideox}, VideoCrafter2~\cite{chen2024videocrafter2} and OpenSora~\cite{opensora}), getting 1,795 annotated video pairs with preference.

\begin{figure}[h]
    \centering
    \includegraphics[width=0.9\linewidth]{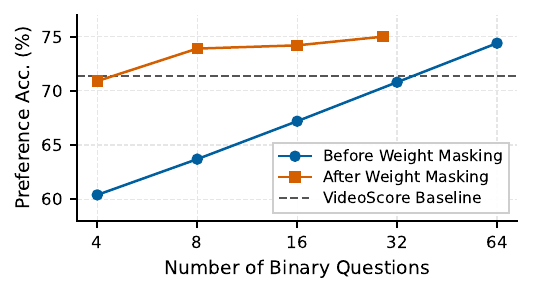}
    \caption{The accuracy of \model on GenAI-Bench improves as the number of binary questions increases. After masking weights from full regression, \model maintains high performance.}
    \label{fig:accuracy_vs_dimension}
\end{figure}

\begin{figure}[h]
    \centering
    \vspace{-0.05cm}
    \includegraphics[width=\linewidth]{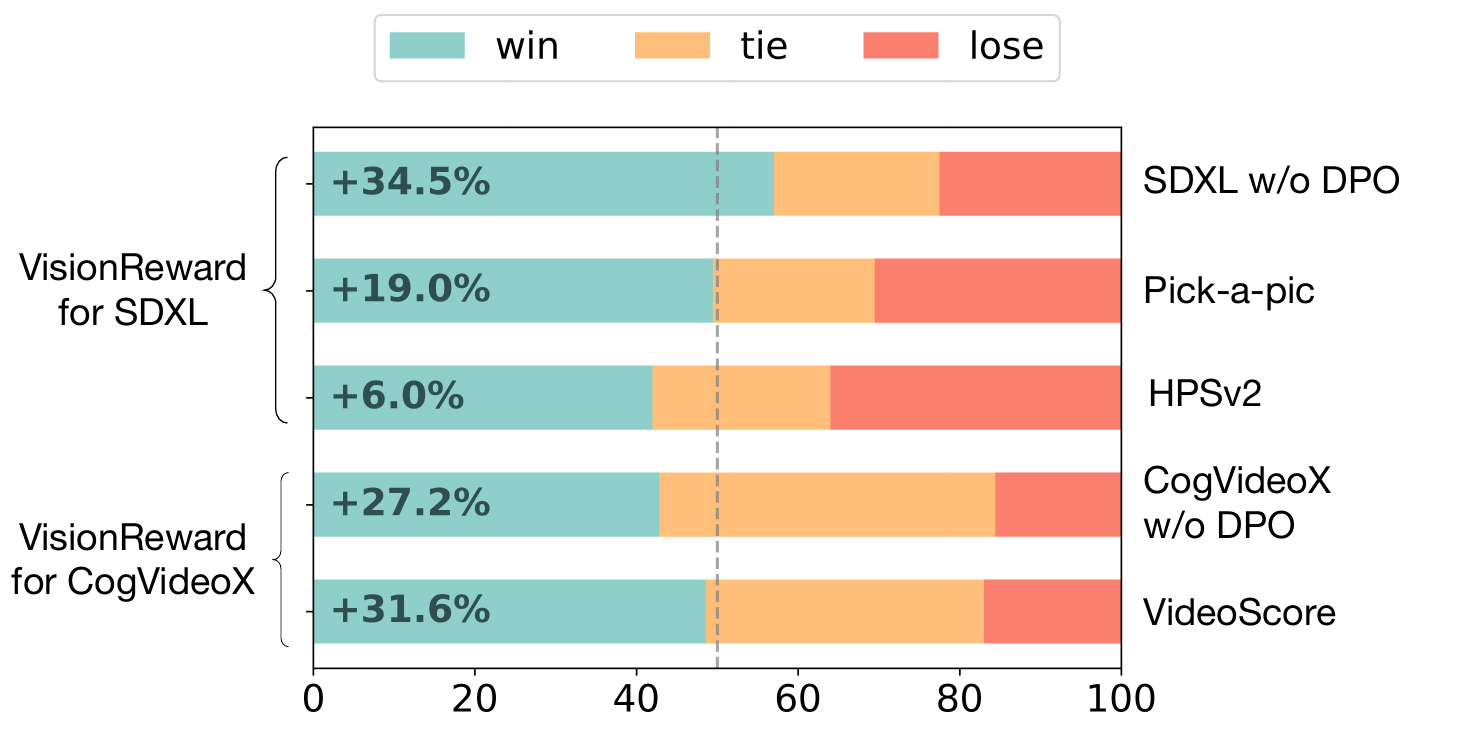}
    \caption{Human evaluation results of DPO using different reward models or preference datasets. We require five annotators to comprehensively evaluate two samples and select the better one. \model achieve the best performance.}
    \label{fig:MPO_human}
\end{figure}

To establish a comprehensive evaluation benchmark for both image and video generation, we construct \textbf{\bench}, which contains separate test sets for images and videos, each consisting of 1,000 prompts.
More details are introduced in Appendix~\cref{sec:X_3_bench}.

\vpara{Main Results: Preference Accuracy.}
Preference accuracy means the probability that a reward model has the same judgment as humans about which image is better. We use \bench to construct our test set for human preference, using SDXL~\cite{podell2023sdxl} to generate images and CogVideoX~\cite{yang2024cogvideox} / VideoCrafter2~\cite{chen2024videocrafter2} / OpenSora~\cite{opensora} to generate videos, resulting in 500 pairs for image and 1,000 pairs for video. We employ annotators to assess the generated images using a preference rating scale from 1 to 5 (with 3 indicating no preference). The average preference score is used as the final preference label. We also take HPDv2~\cite{wu2023human} and GenAI-Bench~\cite{jiang2024genai} as test set.

~\cref{table:performance_main} shows that \model obtains state-of-the-state results in multiple datasets. Notably, in video evaluation, image reward models demonstrate competitive performance when the video duration is within 2 seconds (GenAI-Bench). However, when \textbf{the video duration reaches 6 seconds (\bench)}, only \model is capable of accurately predicting human preference, being twice (\textbf{22.1\% over random}) as high as the best (12.5\% over random) among other methods. This indicates that dynamic information in longer videos poses a challenge for RMs, while \model can effectively address this issue after fine-grained visual learning.

\vpara{Ablation Study: Scalability of Question Scale.}
~\cref{fig:accuracy_vs_dimension} shows that accuracy of preference prediction exhibits significant improvement as the question scale increases, and masking minimal weights maintains performance (Details in Appendix ~\cref{sec:X_2_reward}).
Scalability validates the effects of decomposing preferences via fine-grained questions.
We do more analysis of \model in Appendix ~\cref{sec:X_2_reward} and analysis of fine-grained results in ~\cref{sec:X_4_fine_grained_analysis}.




\subsection{\model for Preference Optimization}

To evaluate the efficacy of \model in preference optimization for visual generative systems, we conduct a series of comparative experiments against current state-of-the-art reward models and established preference datasets. 
We use SDXL as text-to-image base model and CogVideoX as text-to-video base model. 
The empirical results presented in ~\cref{fig:MPO_human} demonstrate \model's superior performance, achieving statistically significant improvements in human preference metrics over competing approaches. 

This section focuses on preference optimization for text-to-video using different reward models. Details of text-to-image are provided in Appendix~\cref{subsec:X_mpo_image_exp}.

\vpara{Dataset \& Training Settings.}
For our backbone model, we select CogVideoX-2B. 
The training prompts are sampled from VidProM \cite{wang2024vidprom} (details in Appendix~\cref{sec:X_mpo_prompt}). To adapt these prompts for video generation, we have optimized them following guidelines from CogVideoX \cite{yang2024cogvideox}, which results in roughly 22,000 samples. We generate 4 videos for each prompt, and use \model to score these videos and apply the MPO strategy to select approximately 9,400 effective preference pairs.
In all our experiments, we maintain a batch size of 32, a learning rate of 5e-6, and employ 100 warmup steps followed by linear decay. We set the DPO parameter $\beta$ to 500. The MPO training process spans around 500 steps, equivalent to about 2 epochs.
During training, we save a checkpoint every 40 steps and use a validation set split from the training set to pick the checkpoint with the highest reward.

\vpara{Evaluation Settings.}
To comprehensively assess the MPO models, we have conducted both automatic and human evaluations. The automatic evaluation is conducted across various benchmarks, including VBench \cite{huang2024vbench} and our Video-MonetBench.
For VBench, we focus on commonly reported key metrics, including \textit{Human Action}, \textit{Scene}, \textit{Multiple Objects}, and \textit{Appearance Style}.
In all these experiments, we utilize prompt optimization recommended in CogVideoX.
Our baseline comparisons include the original CogVideoX-2B and DPO with VideoScore.

\begin{table}[h]
  \resizebox{\columnwidth}{!}{
      \centering
      \setlength{\tabcolsep}{8.5pt}
      {
        \begin{tabular}{l|cccc}
        \toprule
            Methods & \Centerstack{Human \\ Action} & Scene & \Centerstack{Multiple \\ Objects} & \Centerstack{Appear. \\ Style}   \\ 
            \midrule
            Original & 98.20 & 55.60 & 68.43 & \textbf{24.20}  \\
            \midrule
            VideoScore & 97.60 & 56.25 & 68.66 & 23.96 \\
            \model &  \textbf{98.40} & \textbf{57.57} & \textbf{71.54} & 24.02 \\
            \bottomrule
        \end{tabular}
      }
 }
   \caption{Evaluation results on VBench.}
    \label{table: video mpo results}
\end{table}

\vpara{Experimental Results.}
The main results are shown in ~\cref{table: video mpo results}. 
When compared to the original CogVideoX-2B, optimization with \model significantly enhances model performance across these benchmarks. In contrast, optimization with VideoScore tends to degrade performance. The empirical evidence substantiates \model's advanced capacity for multi-dimensional optimization. (Case study in Appendix~\cref{subsec:X_mpo_more_results}.)

\subsection{Ablation Study of MPO}
\label{sec:exp_mpo}

\hhide{
We conduct experiments based on CogVideoX-5B. Specifically, we employ three different strategies, setting the threshold of the total score to 0.6, 0.5, and 0.4 respectively, ensuring that the number of pairs obtained through all three strategies is 5k. We use a batch size of 64, a learning rate of 2e-6, the DPO parameter $\beta$ of 500, and the training steps of 300.

To comprehensively illustrate how MPO addresses the factor optimization bias inherent in DPO, we extend the previously described experimental setup by utilizing the total score of \model to select pairs, setting the threshold at 0.8 to ensure consistency in numbers. }

To comprehensively illustrate how MPO addresses the factor optimization bias inherent in DPO, We conduct experiments based on CogVideoX-5B. We set the threshold of the total score to 0.8 for DPO and 0.6 for MPO, ensuring that the number of pairs obtained through all three strategies is 5k. We use a batch size of 64, a learning rate of 2e-6, the DPO parameter $\beta$ of 500, and the training steps of 300. \model is employed to evaluate scores across various dimensions, with the detailed results presented in ~\cref{tab:ablation_mpo_video}. Through the implementation of the MPO strategy, CogVideoX is optimized in such a manner that it avoids the degradation of certain factors (e.g., alignment), thereby achieving improved trade-offs, such as maintaining good preservation while avoiding excessively slow dynamic changes. The empirical evidence further substantiates \model's capacity for algorithm-agnostic preference alignment, as evidenced by comparative testing with other approaches like MaPO~\cite{hong2024margin}.
\begin{table}[h]
\centering
\setlength{\tabcolsep}{2pt}
\resizebox{\columnwidth}{!}{
    \small{
        \begin{tabular}{@{}l|ccccc|c@{}}
        \toprule
        \textbf{Method} & \textbf{Align.} & \textbf{Quality} & \textbf{Dynamic} & \textbf{Physics} & \textbf{Preserv.} & \textbf{Overall} \\ 
        \midrule
        Original & 1.733 & 0.660 & 0.053 & 0.344 & 0.653 & 4.303 \\ 
        \midrule
        DPO       & \color{dt}{1.697} & 0.680 & {0.034} & 0.356 & \textbf{0.741} & 4.515 \\ 
        DPO w/ MPO & \textbf{1.766} & \textbf{0.688} & {0.042} & 0.356 & 0.721 & \textbf{4.573} \\ 
        \midrule
        MaPO       & 1.736 & 0.660 & {0.052} & 0.345 & {0.645} & {4.295} \\ 
        MaPO w/ MPO & \textbf{1.737} & {0.656} & \textbf{0.055} & \textbf{0.349} & {0.649} & \textbf{4.321} \\ 
        \bottomrule
        \end{tabular}
    }
}

\caption{Ablation Study of MPO strategy. Scores are given by \model on \bench.}

\label{tab:ablation_mpo_video}
\end{table}

For efficiency discussion, the experimental results in ~\cref{table:mpo_ablation_number} reveal the comparative effectiveness of the MPO and DPO methods. We analyze the pairs selected in perspective of ``\( R^i \) dominating \( R^j \)'' mentioned in ~\cref{sec:method_mpo}. These results suggest that MPO outperforms the DPO approach in terms of both efficiency and effectiveness.
These findings highlight promising directions for future research in developing novel optimization algorithms and adapting \model for multi-dimensional optimization.

\begin{table}[h]
  \resizebox{\columnwidth}{!}{
      \centering
      \setlength{\tabcolsep}{8pt}
      {
        \begin{tabular}{l|cc|c}
        \toprule
            \textbf{Method} & \textbf{\#Dom.} & \textbf{\#Not-Dom.} & \textbf{Reward} \\
        \midrule
            Original & - & - & 4.303 \\
            DPO w/o MPO & 3814 & 1456 & 4.515 (+0.212) \\
        \midrule
            DPO w/ MPO  & 5028 & 0 & 4.573 (+0.270) \\
            $\Delta$ (\textit{vs} w/o MPO) & +31.8\% & -100\% & +27.4\% \\
        \bottomrule
        \end{tabular}
      }
 }
   \caption{Comparison of MPO and DPO on \bench. ``\#Dom.'' means the number of pairs that match the rule of ``\( R^i \) dominating \( R^j \)'', while ``\#Not-Dom.'' pairs not match.}
  \label{table:mpo_ablation_number}
\end{table}

\hhide{
\begin{table}[h]
  \resizebox{\columnwidth}{!}{
      \centering
      \setlength{\tabcolsep}{4pt}
      {
        \begin{tabular}{l|cccc}
        \toprule
            Methods & Stability & Dynamic & Physics & Preservation   \\ 
            \midrule
            Original & 0.272 & \textbf{0.047} & 0.323 & 0.584  \\
            \midrule
            VideoScore & 0.242 & 0.046 & 0.319 & 0.557 \\
            \model &  \textbf{0.309} & 0.036 & \textbf{0.337} & \textbf{0.661} \\
            \bottomrule
        \end{tabular}
      }
 }
   \vspace{-0.3cm}
   \caption{Evaluation results on MonetBench.}
    \label{table: video mpo monetbench results}
    \vspace{-3mm}
\end{table}
}



\hhide{
\vpara{Ablation Study: MPO or Single RM.}
To evaluate the effectiveness of multi-objective versus single-objective direct preference learning, we utilize \model and conduct a comparative experiment using a weighted single score as the DPO. Specifically, we sample 760k pairs under the MPO method, and sample 780k pairs using the weighted single score (threshold of 0.014). We train for 2.2k iterations with a batch size of 1,024 and a learning rate of 1e-8. The results are shown in ~\cref{table:X_mpo_ablation1}, which indicates that multi-objective can achieve better results.
\begin{table}[h]
  \resizebox{\columnwidth}{!}{
      \centering
      {
        \begin{tabular}{l|cccc}
        \toprule
            Methods & CLIP & Aes & HPSv2 & PickScore \\
        \midrule
            Baseline & 0.273 & 5.463 & 0.282 & 22.25 \\
            DPO with \model & \textbf{0.278} & 5.664 & \textbf{0.291} & 22.227 \\
            MPO with \model & \textbf{0.278} & \textbf{5.719} & \textbf{0.291} & \textbf{22.505} \\
        \bottomrule
        \end{tabular}
      }
 }
   \vspace{-0.2cm}
   \caption{Evaluation results on DrawBench.}
    \vspace{-3mm}
  \label{table:X_mpo_ablation1}
\end{table}


\vpara{Ablation Study: MPO or Single RM.}
To comprehensively illustrate how MPO addresses the factor optimization bias inherent in DPO, we extend the previously described experimental setup by utilizing the total score of \model to select pairs, setting the threshold at 0.8 to ensure consistency in numbers. \model is employed to evaluate scores across various dimensions, with the detailed results presented in ~\cref{table:ablation_video_mpo_dpo}. Through the implementation of the MPO strategy, CogVideoX is optimized in such a manner that it avoids the degradation of certain factors (e.g., alignment), thereby achieving improved trade-offs, such as maintaining good preservation while avoiding excessively slow dynamic changes.

\begin{table*}[t]
  \resizebox{\textwidth}{!}{%
  \centering
  \renewcommand{\arraystretch}{1.15}
  \setlength{\tabcolsep}{10pt}
  \ 
  {
    \begin{tabular}{c|cccccccc|c}
    \toprule
    Method & Alignment & Composition & Quality & Fidelity & Stability & Dynamic & Physics & Preservation & Total \\
    \addlinespace[-0.1mm]
        \midrule
        Baseline & 1.733 & 0.048 & 0.660 & 0.497 & 0.315 & \textbf{0.053} & 0.344 & 0.653 & 4.303 \\
        DPO & \textcolor{red}{1.697} & \textbf{0.053} & 0.680 & \textbf{0.591} & \textbf{0.363} & \textbf{\textcolor{red}{0.034}} & \textbf{0.356} & \textbf{0.741} & 4.515 \\
        MPO & \textbf{1.766} & \textbf{0.053} & \textbf{0.688} & 0.587 & 0.360 & \textcolor{red}{0.042} & \textbf{0.356} & 0.721 & \textbf{4.573} \\
    \bottomrule
    \end{tabular}
  }
 }
   \vspace{-0.2cm}

   \caption{Evaluation results of MPO, DPO, and baseline. Scores are given by \model on \bench.}
    \vspace{-3mm}

  \label{table:ablation_video_mpo_dpo}
\end{table*}

}



\hhide{
\vpara{Ablation Study: Different Strategy for MPO.}
In using the MPO algorithm, we define the way in which \( R^i \) dominates \( R^j \) (given two images \( x^i \) and \( x^j \)) to select pairs. The definition of ``dominate'' includes at least three methods for different reward objective: score of each dimension, score of each sub-dimension, and score of each check question. To investigate the impact of different definitions of ``dominate'', we conduct experiments based on CogVideoX-5b. Specifically, we employ three different strategies, setting the threshold of the total score to 0.6, 0.5, and 0.4 respectively, ensuring that the number of pairs obtained through all three strategies is 5k. We use a batch size of 64, a learning rate of 2e-6, the DPO parameter $\beta$ of 500, and the training steps of 300. After training, we compare the evaluation results of \model on \bench. ~\cref{tab:ablation_mpo_strategy} shows that using the score for each dimension as the reward objective yields the best results.

\begin{table}[h]
    \resizebox{\columnwidth}{!}{
        \centering
        \setlength{\tabcolsep}{5pt}
        \begin{tabular}{@{}l|cccc}
        \toprule
        Methods & Baseline & Dimension & Sub-dimension & Question \\
        \midrule
        Reward & 4.303 & \textbf{4.573} & 4.539 & 4.514 \\
        \bottomrule
        \end{tabular}
    }
    \vspace{-0.2cm}
    \caption{Score of \model after different strategies of MPO. Dimension: ``dominate'' based on score of each dimension. Sub-dimension: ``dominate'' based on score of each sub-dimension. Question: ``dominate'' based on score of each binary question.}
    \vspace{-3mm}
    \label{tab:ablation_mpo_strategy}
\end{table}
}

\hhide{
\vpara{Details of Prompt Used in VLM / LLM.}
In the preference experiment, we compare our model with GPT-4o~\cite{achiam2023gpt} and Gemini~\cite{team2024gemini}. We follow the prompt template from the VideoScore~\cite{he2024videoscore} evaluation, requesting that the large models provide assessments and scores across several dimensions, and use these scores to determine the accuracy of preferences. For images, we adopt a similar approach, defining dimensions in the prompt consistent with the annotation process, and determining preferences based on the relative scores.
}

\hhide{
\vpara{Main Results: Accuracy on Judgment.}
To evaluate the effectiveness of judgment learning, we construct a visual quality QA set to assess the visual assessment capabilities of \model compared to other VLMs. We collect 1,364 test cases for images involving 14 types of questions across 4 dimensions, and additionally 1,308 cases covering 8 types of questions across 4 dimensions related to \textbf{dynamic content in videos}. To ensure the generality of these questions, we combine adjacent degrees in the checklists under each sub-dimension, enhancing distinctiveness and minimizing incidental subjectivity. ~\cref{table:meta_qa_vlm} demonstrates \model's superiority in judging visual quality over existing generalized multimodal LLMs, which remains a challenge for generalized multimodal LLMs.
}

\section{Conclusion}

We introduce \model, a reward model for visual generation, which is fine-grained and multi-dimensional. By enabling Vision-Language Model (VLM) to perform binary assessments and applying linear summation with weighting coefficients derived from preference learning, \model achieves highly accurate and interpretable. For visual generative optimization, \model surpasses other reward models and enable multi-dimensional strategy.

\section*{Acknowledgments}
This research was supported by Natural Science Foundation of China (NSFC) No. 62276148, NSFC No. 62495063. The authors would like to thank Z.AI for sponsoring the computation resources used in this work.


\bibliography{main}



\clearpage

\twocolumn[
        \centering
        \Large
        \textbf{Appendix}\\
        \vspace{1.0em}
       ]

\section{More Details of Annotation}
\label{sec:X_1_anno}

\subsection{Details of Annotation Design}
\label{sec:X_1_anno_design_detail}

To ensure the diversity of our annotation, we collect our annotated data from various sources, as presented in ~\cref{tab:anno_source}. 
The prompt and example for data cleaning is shown in ~\cref{tab:X_prompt_video_clean}.
~\cref{tab:checklist_tag} illustrates our preference dimensions and the checklist count. We identify 5 dimensions for text-to-image generation and expand to 9 dimensions for text-to-video.

\begin{table}[h]
    \centering
    
    \resizebox{\columnwidth}{!}{
        \small{
            \begin{tabular}{c|ccc}
            \toprule
            \textbf{Type} & \textbf{Source} & \textbf{\#Samples} & \textbf{\#Checklist} \\
            \midrule
            \multirow{3}{*}{Image} & ImageRewardDB~\cite{xu2023imagereward} & 16K & 1M \\
             & Pick-a-Pic~\cite{kirstain2023pick} & 16K & 1M \\
             & HPDv2~\cite{wu2023human} & 16K & 1M \\
            \midrule
            \multirow{4}{*}{Video} & CogVideoX~\cite{yang2024cogvideox} & 10K & 0.6M \\
             & Open-Sora~\cite{opensora} & 10K & 0.6M \\
             & VideoCrafter2~\cite{chen2024videocrafter2} & 10K & 0.6M \\
             & Panda-70M~\cite{chen2024panda} & 3K & 0.2M \\
            \bottomrule
            \end{tabular}
        }
    }
    \vspace{-0.15cm}
    
    \caption{Statistics of source data and annotation.}
    
    \label{tab:anno_source}
\end{table}

\begin{table}[h]
  \centering
  \resizebox{\columnwidth}{!}{
    \small{
        \begin{tabular}{c|cccc}
        \toprule
            \multirow{2}{*}{\textbf{Dimension}} & \multicolumn{2}{c}{\textbf{\#Sub-dimension}} & \multicolumn{2}{c}{\textbf{\#Checklist}}\\
        \cmidrule(r){2-3} \cmidrule(r){4-5}
        & Image & Video & Image & Video \\
        \addlinespace[-0.1mm]
            \midrule
            Alignment & 1 & 1 & 1 & 4 \\
            Composition & 5 & 1 & 13 & 2 \\
            Quality & 5 & 4 & 14 & 14 \\
            Fidelity & 5 & 3 & 25 & 9 \\
            Safety\&Emotion & 2 & 1 & 8 & 4 \\
            \midrule
            Stability & - & 5 & - & 12 \\
            Dynamic & - & 2 & - & 8 \\
            Physics & - & 1 & - & 4 \\
            Preservation & - & 2 & - & 7 \\
            \midrule
            Total & 18 & 20 & 61 & 64 \\
            \bottomrule
        \end{tabular}   
    }
 }
 \vspace{-0.15cm}
 \caption{Taxonomy of annotation for \model.}
 \vspace{-0.5cm}
 \label{tab:checklist_tag}
\end{table}

    
    
    

\begin{figure*}[t]
    \centering
    \begin{subfigure}{0.45\textwidth}
        \includegraphics[width=\linewidth]{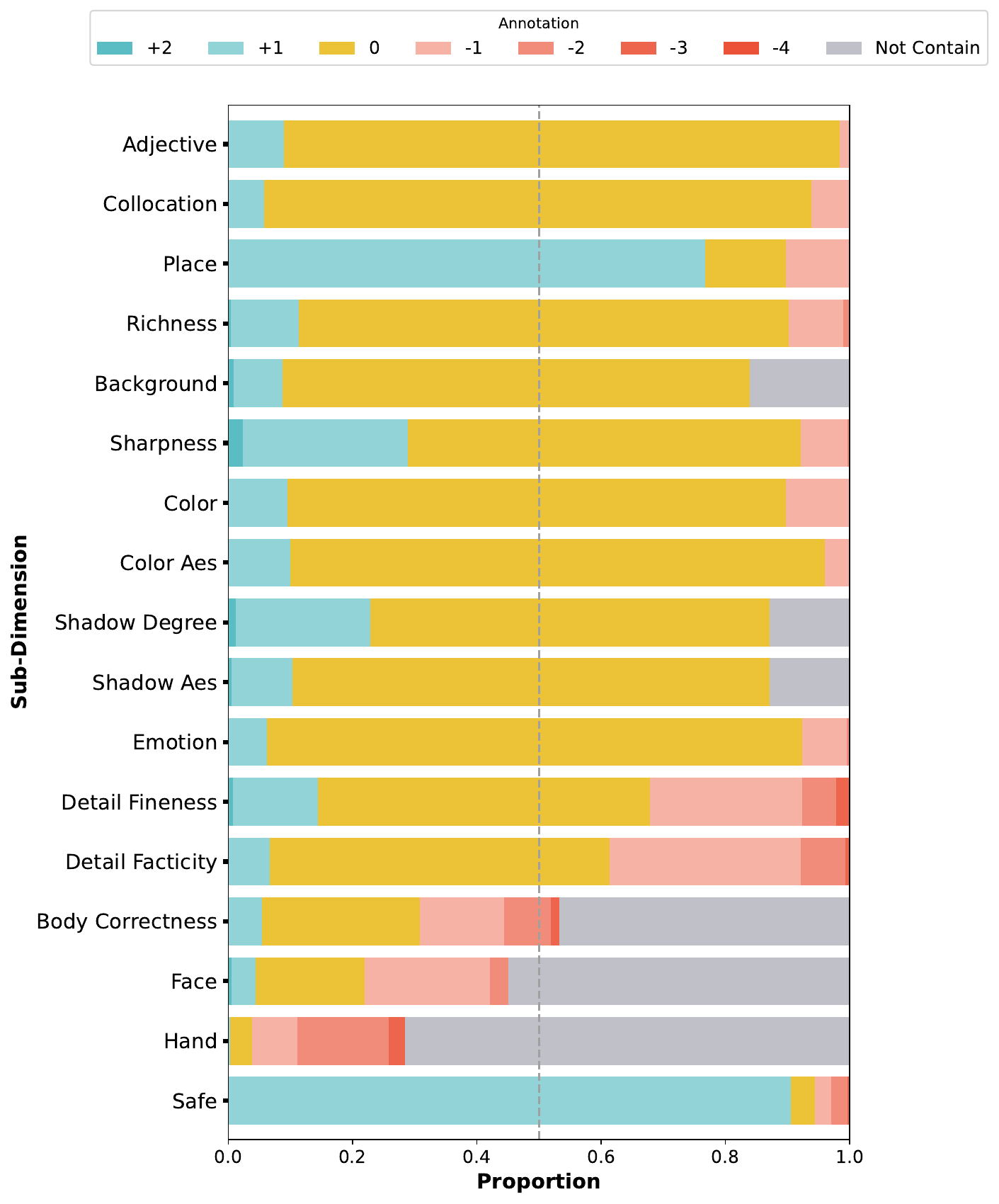}
        \caption{Text-to-image}
    \end{subfigure}
    \begin{subfigure}{0.465\textwidth}
        \includegraphics[width=\linewidth]{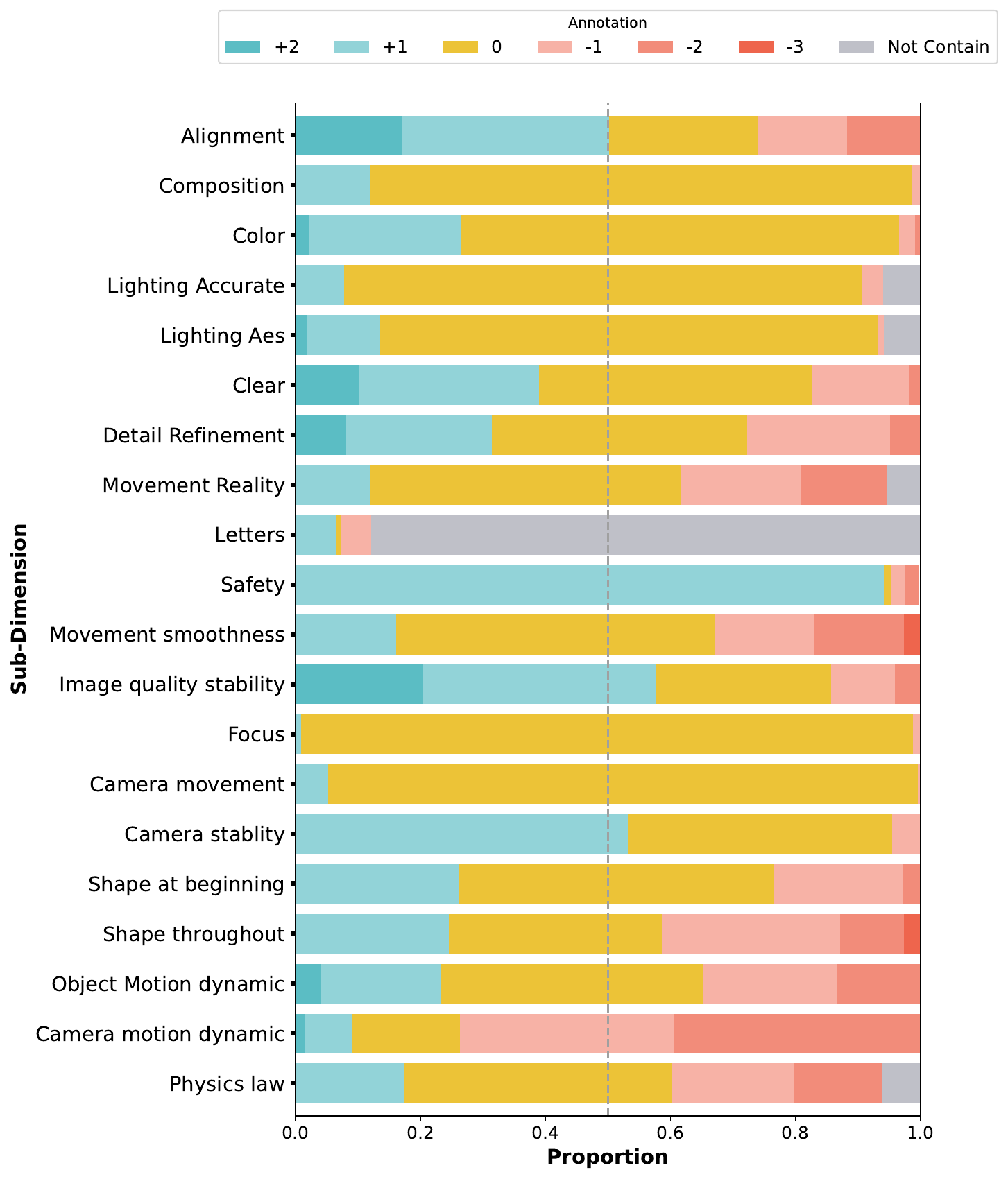}
        \caption{Text-to-video}
    \end{subfigure}
    \caption{Annotation statistics of different sub-dimensions.}
    \label{fig:attributes_annotation_distribution}
\end{figure*}

\newcolumntype{Y}{>{\RaggedRight\arraybackslash}X}

\begin{table}[h]
    \centering
    \begin{tabularx}{\linewidth}{c|Y}
        \toprule
        SYSTEM & Assume you are a model responsible for refining and polishing English expressions. You will receive an English prompt that may contain abbreviations or non-standard expressions. Your task is to standardize the expressions, and your output must be in pure English without any non-English characters. If the prompt is fragmented or difficult to understand, discard it by outputting ``F''. Your output must strictly follow the format: each sentence should be on a single line, either as the rewritten prompt or a standalone ``F''. \\
        \midrule
        USER & Here is the prompt you have received: [[PROMPT]] \\
        \midrule
        INPUT & Soft rays of light through the many different types of trees inside a forest, sunrise, misty,  photorealistic, ground level, -neg \&quot;no large bodies of water\&quot; -ar 16:9   4K, -ar 16:9 \\
        \midrule
        OUTPUT & The soft rays of light filter through the myriad types of trees within the forest at sunrise, creating a misty, photorealistic scene from ground level. Exclude any large bodies of water. The aspect ratio should be 16:9 in 4K resolution. Aspect ratio: 16:9. \\
        \bottomrule
    \end{tabularx}
    \caption{Prompt template and example for prompt cleaning.}
    \label{tab:X_prompt_video_clean}
\end{table}

\subsection{Statistics of Annotation Result}
\label{sec:X_1_anno_stat}

~\cref{fig:attributes_annotation_distribution} is the statistical results of the labeled data for images and videos. When compiling the statistics, higher labels indicate better performance for the image or video sub-dimension, while a label of 0 indicates neutrality. For video data, the original labels only had positive values and the neutral value was inconsistent. Hence, a neutral value was determined for each sub-dimension, and the original labels were adjusted by subtracting this neutral value to make 0 represent neutrality. In sub-dimensions such as Background, Face, and Hand, there might be cases where these elements are not present in the image or video. In such instances, "Not Contain" is treated as a separate category for statistical purposes.

There are two main characteristics to note.
\begin{itemize}
    \item For most sub-dimensions, the distribution of options roughly follows a normal distribution, with the majority being ordinary, and the quantities of instances with extreme characteristics, either very good or very bad, are reduced. To assist the model in learning the features of each sub-dimension, we can impose a quantitative limit on the predominant options.
    \item Certain sub-dimensions, such as the presence of hands, require a mask when predicting human preferences. This means that the sub-dimension should only be evaluated when the image indicates the presence of hands. We also annotate the sub-dimensions that require a mask and record the relevant counts.
\end{itemize}

\section{More Details and Results of \model}
\label{sec:X_2_reward}

\model comprises two steps: visual judgment and linear regression. 

\vpara{Visual Judgment Process.}
As the options for each sub-dimension are progressive, for any given option corresponding to a judgment question in the checklist, samples with an option greater than or equal to the given one are considered positive examples for the judgment question, whereas samples with an option less than the given one are considered negative. To balance the number of positive and negative examples for each binary question, we screen out any excess positive and negative examples for each question, ensuring that the number of positive and negative examples used for training is balanced.
For alignment, we use different methods on images and videos. For images, we use VQAScore~\cite{lin2025evaluating} as an alignment judgment. For videos, we train five levels of judgment for \model.

\vpara{Main Results: Accuracy on Judgment.}
To evaluate the effectiveness of judgment learning, we construct a visual quality QA set to assess the visual assessment capabilities of \model compared to other VLMs. We collect 1,364 test cases for images involving 14 types of questions across 4 dimensions, and additionally 1,308 cases covering 8 types of questions across 4 dimensions related to \textbf{dynamic content in videos}. To ensure the generality of these questions, we combine adjacent degrees in the checklists under each sub-dimension, enhancing distinctiveness and minimizing incidental subjectivity. ~\cref{table:meta_qa_vlm} demonstrates \model's superiority in judging visual quality over existing generalized multimodal LLMs, which remains a challenge for generalized multimodal LLMs.

\begin{table}[h]
    \centering
    
    \resizebox{\columnwidth}{!}{
        \small{
            \setlength{\tabcolsep}{10pt}
            \begin{tabular}{cccc}
            \toprule
            Composition & Quality & Fidelity & Safety \\
            \midrule
            97.9 & 98.2 & 98.3 & 99.1 \\
            \midrule
            Stability & Dynamic & Physics & Preservation \\
            \midrule
            97.4 & 99.9 & 88.2 & 99.8 \\
            \bottomrule
            \end{tabular}
        }
    }
    \vspace{-0.3cm}
    
    \caption{Consistency of \model in each dimension.}
    \vspace{-0.3cm}
    
  \label{table:vlm_consistency}
\end{table}

\vpara{Main Results: Consistency on Judgment.}
As questions corresponding to each sub-dimension assess varying degrees of a particular factor, it's important to measure consistency of \model across multiple questions of the same sub-dimension. Consistency measures the likelihood that the model provides consistent responses across a series of judgments concerning this factor. ~\cref{table:vlm_consistency} shows that \model has high consistency (more than 97 \%) in most (7 of 8) dimensions.

\begin{algorithm}[H]
\footnotesize
    \renewcommand{\algorithmicrequire}{\textbf{Input:}}
    \renewcommand{\algorithmicensure}{\textbf{Output:}}
    \caption{Iterative Regression with Weight Masking}
    \begin{algorithmic}[1]
        \REQUIRE Dataset of human preferences $\mathcal{D} = \{(\mathbf{X}_i, \mathbf{X}_j, y)\}$, where each pair $(\mathbf{X}_i, \mathbf{X}_j)$ represents feature vectors of binary responses and $y \in \{0, 1\}$ is the human preference label
        \STATE \textbf{Initialization:} Initialize linear weights $\mathbf{w} = [w_1, \ldots, w_n]$
        \STATE Initialize convergence criterion $\text{diff} \gets \infty$
        \WHILE{$\text{diff} > \epsilon$}
            \STATE $\textbf{w}_{\text{old}} \gets \textbf{w}$
            \FOR{each $(\mathbf{X}_i, \mathbf{X}_j, y)$ in $\mathcal{D}$}
                \STATE $\Delta \mathbf{X} \gets \mathbf{X}_i - \mathbf{X}_j$
                \STATE $\hat{y} \gets \sigma(\Delta \mathbf{X}^T \mathbf{w})$
                \STATE $\nabla_{\textbf{w}} \text{Loss} = (\hat{y} - y) \Delta \mathbf{X}$
                \STATE $\mathbf{w} \gets \mathbf{w} - \alpha \nabla_{\textbf{w}} \text{Loss}$
            \ENDFOR
            \STATE Mask negative weights $\mathbf{w} \gets \mathbf{w} \odot (\mathbf{w} > 0)$
            \STATE $\text{diff} \gets ||\textbf{w} - \textbf{w}_{\text{old}}||$
        \ENDWHILE
        \ENSURE Trained weights $\mathbf{w}$
    \end{algorithmic}
    \label{iterative_logistic_regression}
\end{algorithm}
\vspace{-5pt}

\begin{table*}[t]
  \resizebox{\textwidth}{!}{%
  \centering
  \renewcommand{\arraystretch}{1.15}
  \setlength{\tabcolsep}{8pt}  
  \ 
  {
    \begin{tabular}{lcccccccc}
    \toprule
        \multirow{2}{*}{\textbf{Method}} & \multicolumn{4}{c}{\textbf{Image}} & \multicolumn{4}{c}{\textbf{Video}}\\
    \cmidrule(r){2-5} \cmidrule(r){6-9}
    & Composition & Quality & Fidelity & Safety & Stability & Dynamic & Physics & Preservation \\
    \addlinespace[-0.1mm]
        \midrule
        LLaVa$^*$ & \text{59.9} & \text{65.7} & \text{59.8} & \text{64.4} & \text{52.5} & \text{53.8} & \text{50.6} & \text{47.5} \\
        CogVLM2~\cite{hong2024cogvlm2} & \text{65.8} & \text{67.1} & \text{53.1} & \text{74.7} & \text{49.3} & \text{57.1} & \text{51.2} & \text{47.8} \\
        GPT-4o~\cite{achiam2023gpt} & \text{73.1} & \text{62.7} & \text{61.9} & \text{70.1} & \text{57.9} & \text{69.1} & \text{62.4} & \text{58.8} \\
        Gemini~\cite{team2024gemini} & \text{69.4} & \text{59.9} & \text{59.7} & \text{74.9} & \text{58.1} & \text{71.1} & \text{58.1} & \text{59.6} \\
        \model (Ours) & \textbf{78.8} & \textbf{81.1} & \textbf{80.9} & \textbf{83.9} & \textbf{64.8} & \textbf{75.4} & \textbf{68.1} & \textbf{72.0} \\
    \bottomrule
    \end{tabular}
  }
 }
   \vspace{-0.2cm}

   \caption{Accuracy of \model and other vision-language models (VLMs) on vision quality questions constructed from our annotation. $^*$We test LLaVA-v1.5-7B~\cite{liu2023llava} for image and LLava-Next-Video-34B~\cite{li2024llava} for video. }

  \label{table:meta_qa_vlm}
\end{table*}

\vpara{Weight Masking.}
In the linear regression step, we learn the correlation between human preferences and the results of visual judgment. In our design, if the result of the visual judgment is ``yes'', human preference improves. We examine the correlation between human preference and each judgment result, and the numerical results indicate a positive correlation. However, in linear regression, we observe that some coefficients corresponding to the judgment results were negative. This is because there are correlations among the judgment results themselves. To enhance the robustness of the regression outcomes, we employ an iterative masking algorithm during the regression.


\vpara{Ablation Study: Impact of Training Set Size.}
We conduct experiments with varying sizes of the training set to investigate its influence on regression performance. As demonstrated in ~\cref{tab:X_regression_exp}, the accuracy improves monotonically as the training set size increases up to 4,000 samples, beyond which the accuracy stabilizes.
\begin{table}[h]
    \resizebox{\columnwidth}{!}{
        \centering
        \setlength{\tabcolsep}{10pt}
        \begin{tabular}{l|cccccc}
        \toprule
        \textbf{Size} & 200 & 500 & 1k & 2k & 4k & 8k \\
        \midrule
        \textbf{Acc.} & 77.6 & 80.3 & 80.6 & 80.9 & 81.3 & 81.3 \\
        \bottomrule
        \end{tabular}
    }
    \vspace{-0.2cm}
    \caption{Accuracy on HPDv2-test for different sizes of train set.}
    \vspace{-1mm}
    \label{tab:X_regression_exp}
\end{table}

\hhide{
\begin{table}[h]
    \resizebox{\columnwidth}{!}{
        \centering
        \setlength{\tabcolsep}{15pt}
        \begin{tabular}{l|cccc}
        \toprule
        \textbf{Size} & 100 & 200 & 500 & 1k \\
        \midrule
        \textbf{Accuracy} & 76.5 & 77.6 & 80.3 & 80.6 \\
        \midrule
        \textbf{Size} & 2k & 4k & 8k & 16k \\
        \midrule
        \textbf{Accuracy} & 80.9 & 81.3 & 81.2 & 81.3 \\
        \bottomrule
        \end{tabular}
    }
    \vspace{-0.2cm}
    \caption{Average accuracy on HPDv2 test set for different sizes of regression train set.}
    \vspace{-3mm}
    \label{tab:X_regression_exp}
\end{table}
}


\section{More Details and Results of MPO}

\subsection{Details of Prompt for MPO}
\label{sec:X_mpo_prompt}
We employ the prompt filtering approach proposed by UniFL~\cite{zhang2024uniflimprovestablediffusion} to curate our dataset. This strategy comprises two pivotal steps:  
\begin{itemize}  
    \item \textbf{Semantic-Based Filtering:}   
    Utilizing an existing scene graph parser~\cite{8954449}, we evaluate the semantic richness of prompts by analyzing the number of subjective and objective relationships. Prompts with fewer than one meaningful relationship are filtered out to reduce noise data.
    \item \textbf{Cosine Similarity-Based Selection:}   
    Following the semantic filtering process, we apply a cosine similarity-based iterative selection mechanism. By maintaining a maximum similarity threshold of 0.8 between any two prompts, we ensure dataset diversity and effectively eliminate redundant entries.  
\end{itemize}  


\subsection{\model for Text-to-Image Optimization}
\label{subsec:X_mpo_image_exp}

\vpara{Dataset \& Training Settings.}
We strategically sample 63,165 prompts from existing datasets and generate 8 images per prompt using SDXL (this procedure theoretically produces 1.76M text-image pairs). Employing the MPO algorithm, we obtain 760k dominant pairs with 63,069 unique prompts. For comparison, we use HPSv2 with a threshold of 0.0015, getting 770k pairs with 63,107 unique prompts from the same source. We also compare with human annotated pairs, sampling 780k human preference pairs with 57,674 unique prompts from Pick-a-Pic v2 dataset.

We maintain consistent training parameters and dataset sizes across all experiments to ensure fair comparison. For all three experiments, we used an effective batch size of 256 (with GAS set to 4 and train batch size set to 1), set $\beta$ to 5000, and a learning rate of 5e-9 (before scaling). 
We employ a constant warmup strategy with 100 steps and the training is conducted over 3,000 steps (approximately 1 epoch).

\vpara{Evaluation Settings.}
We conducted both automatic and human evaluation on DrawBench~\cite{saharia2022photorealistic}. Automatic evaluation includes multiple metrics such as human preference RMs, CLIP~\cite{radford2021learning} and LAION-Aesthetic~\cite{schuhmann2022laion}. 


\vpara{Experimental Results.}
Main results are demonstrated in ~\cref{table:mpo_image_score} and ~\cref{table:mpo_image_analysis}. 
\model gets leading results across multiple machine metrics and achieves significant improvements across all four dimensions of \model.

\begin{table}[h]
  \resizebox{\columnwidth}{!}{
      \centering
      {
        \begin{tabular}{l|cccc}
        \toprule
            Methods & CLIP & Aes & HPSv2 & PickScore \\
        \midrule
            Original & 0.273 & 5.463 & 0.282 & 22.25 \\
            \midrule
            Pick-a-Pic & \textbf{0.279} & 5.511 & 0.286 & 22.45 \\
            HPSv2 & 0.277 & 5.599 & \textbf{0.292} & 22.58 \\
            \model & \textbf{0.279} & \textbf{5.612} & 0.289 & \textbf{22.61} \\
        \bottomrule
        \end{tabular}
      }
 }
   \vspace{-0.3cm}
   \caption{Evaluation results of multiple metrics on DrawBench.}
    \vspace{-3mm}
  \label{table:mpo_image_score}
\end{table}

\begin{table}[h]
  \resizebox{\columnwidth}{!}{
      \centering
      \setlength{\tabcolsep}{5pt}
      {
        \begin{tabular}{l|cccc}
        \toprule
            Methods & Composition & Quality & Fidelity & Safety \\
        \midrule
            Original & 0.755 & 0.550 & 0.009 & -0.008 \\
            \midrule
            Pick-a-Pic  & 0.765 & 0.588 & 0.009 & -0.009 \\
            HPSv2 & 0.874 & 0.630 & 0.010 & -0.004 \\
            \model & \textbf{0.894} & \textbf{0.670} & \textbf{0.017} & \textbf{-0.001} \\
        \bottomrule
        \end{tabular}
      }
 }
   \vspace{-0.3cm}
   \caption{Evaluation results on \bench.}
    \vspace{-3mm}
  \label{table:mpo_image_analysis}
\end{table}



\subsection{More Results of MPO}
\label{subsec:X_mpo_more_results}

\begin{figure*}
    \centering
    \includegraphics[width=\linewidth]{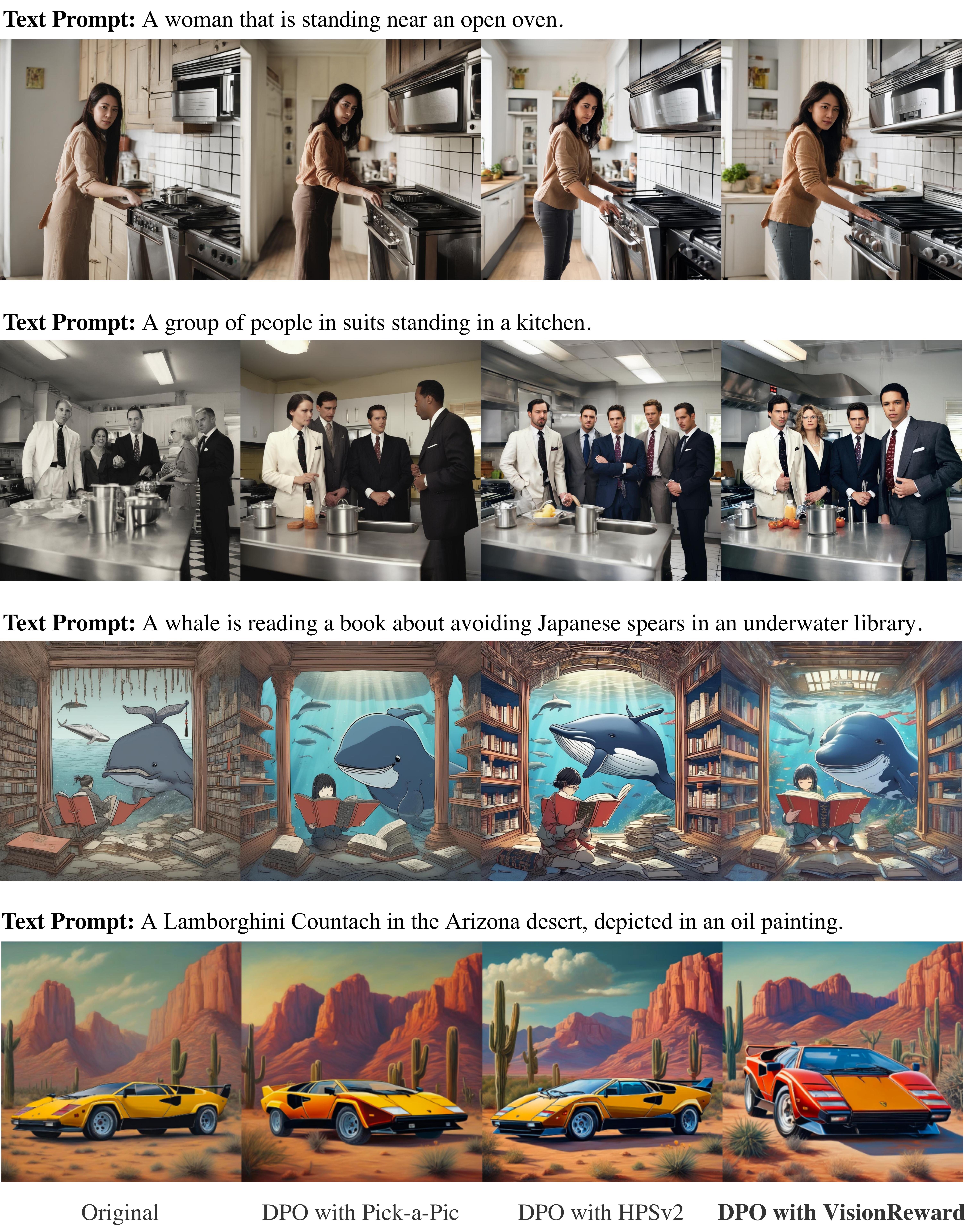}
    \caption{Qualitative result of MPO in text-to-image.}
    \label{fig:mpo_demo_image_1}
\end{figure*}

\begin{figure*}
    \centering
    \vspace{0.2cm}
    \includegraphics[width=\linewidth]{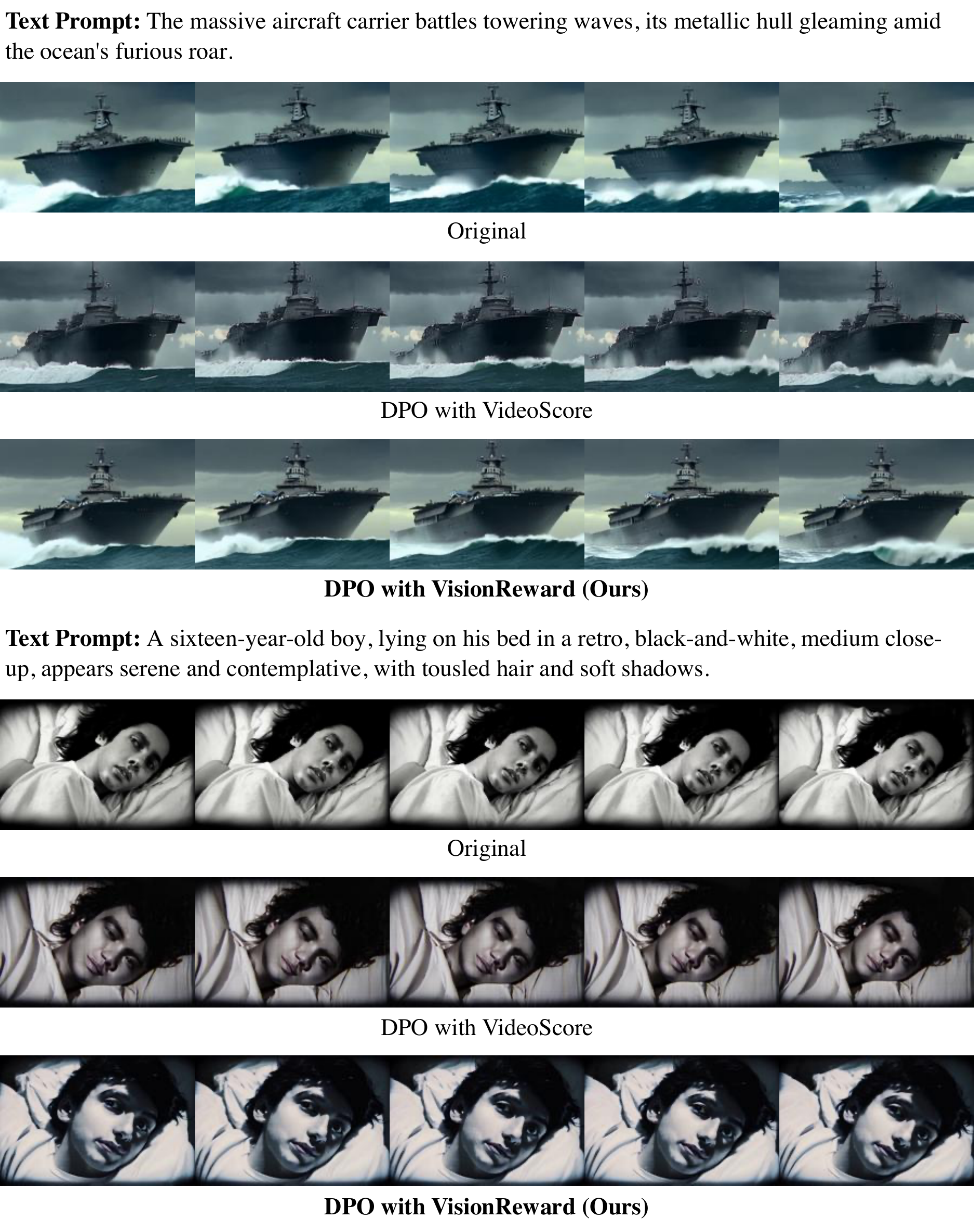}
    \caption{Qualitative result of MPO in text-to-video.}
    \label{fig:mpo_demo_video_1}
    \vspace{-0.5cm}
\end{figure*}

\begin{figure*}
    \centering
    \vspace{0.2cm}
    \includegraphics[width=\linewidth]{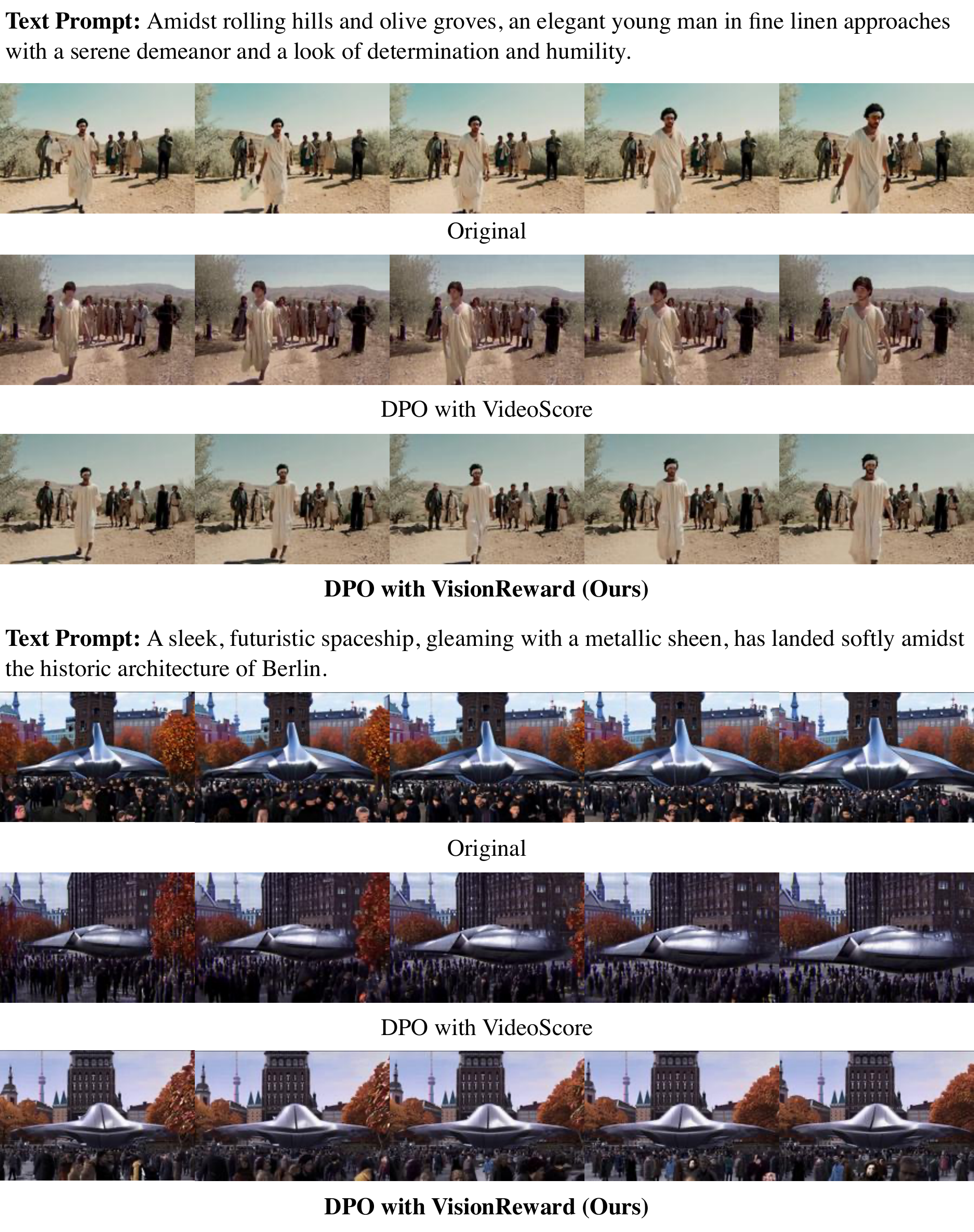}
    \caption{Qualitative result of MPO in text-to-video.}
    \label{fig:mpo_demo_video_2}
    \vspace{-0.5cm}
\end{figure*}

\vpara{Case Study.}
~\cref{fig:mpo_demo_image_1} shows MPO cases for text-to-image, while ~\cref{fig:mpo_demo_video_1} and ~\cref{fig:mpo_demo_video_2} show MPO cases for text-to-video. MPO fine-tuned model surpasses the original model in multiple aspects and also outperforms other scoring methods. 

\begin{figure*}
    \vspace{0.5cm}
    \centering
    \begin{subfigure}{0.45\textwidth}
        \includegraphics[width=\linewidth]{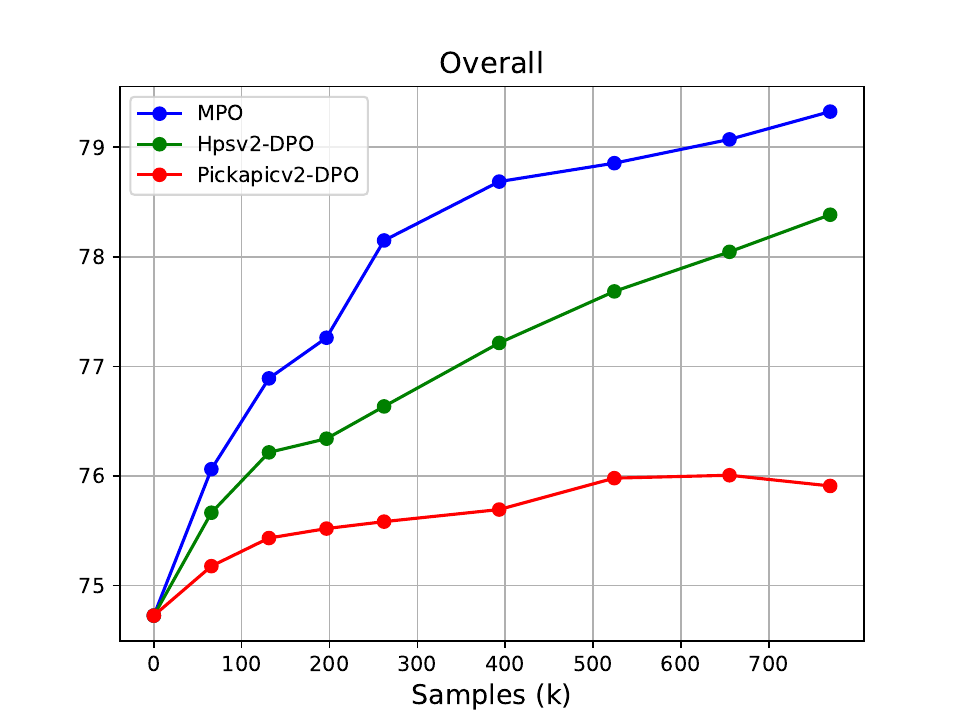}
        \caption{Overall Score}
    \end{subfigure}
    \begin{subfigure}{0.45\textwidth}
        \includegraphics[width=\linewidth]{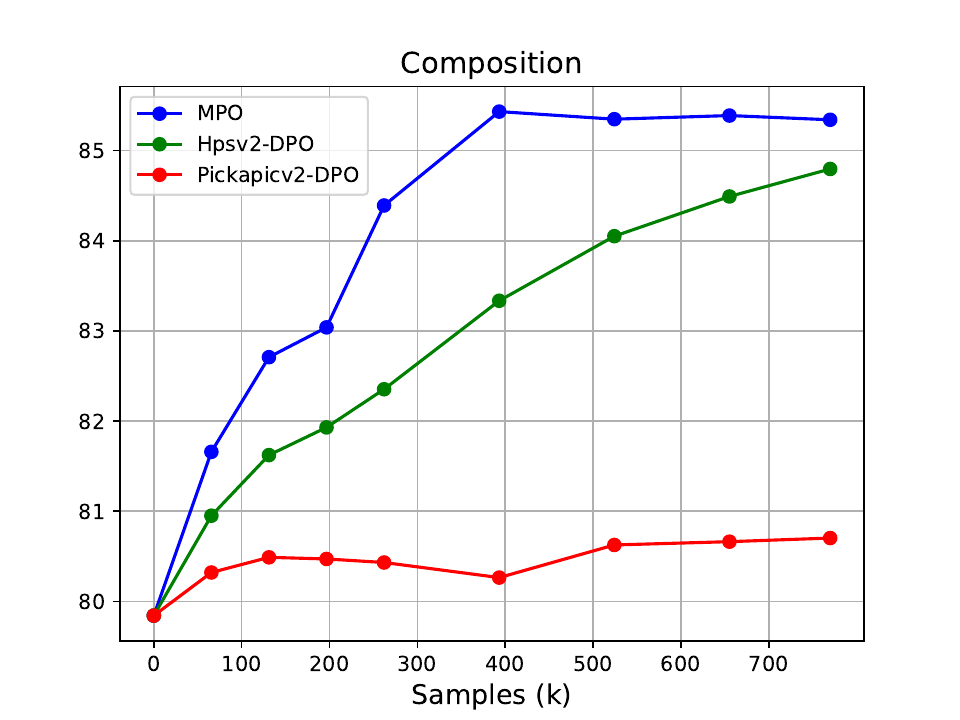}
        \caption{Composition Score}
    \end{subfigure}
    \begin{subfigure}{0.45\textwidth}
        \includegraphics[width=\linewidth]{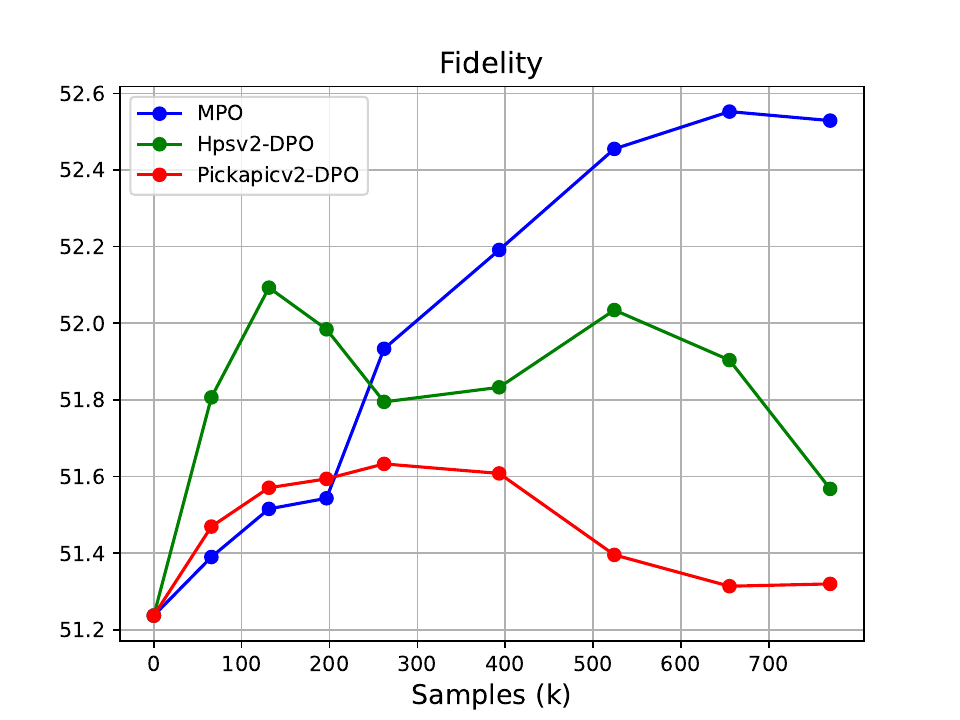}
        \caption{Fidelity Score}
    \end{subfigure}
    \begin{subfigure}{0.45\textwidth}
        \includegraphics[width=\linewidth]{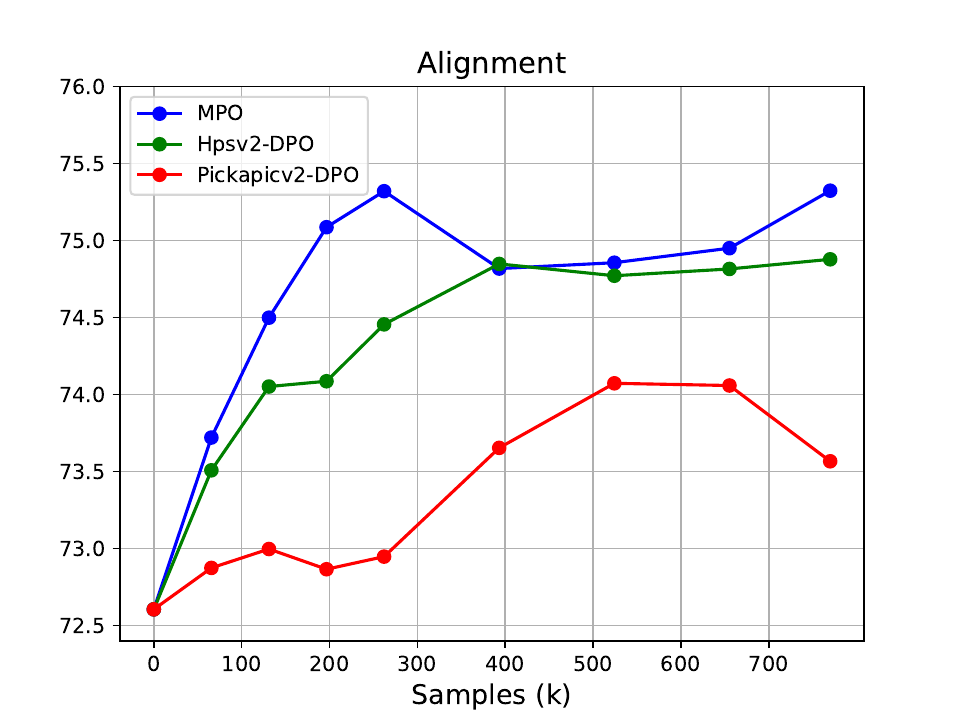}
        \caption{Alignment Score}
    \end{subfigure}
    \begin{subfigure}{0.45\textwidth}
        \includegraphics[width=\linewidth]{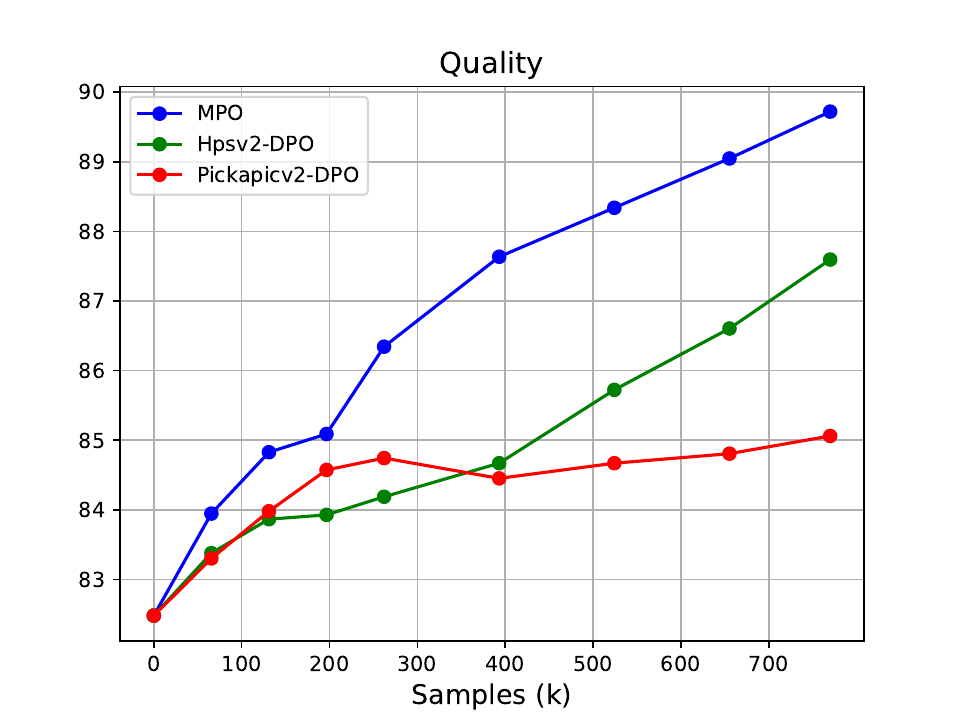}
        \caption{Quality Score}
    \end{subfigure}
    \begin{subfigure}{0.45\textwidth}
        \includegraphics[width=\linewidth]{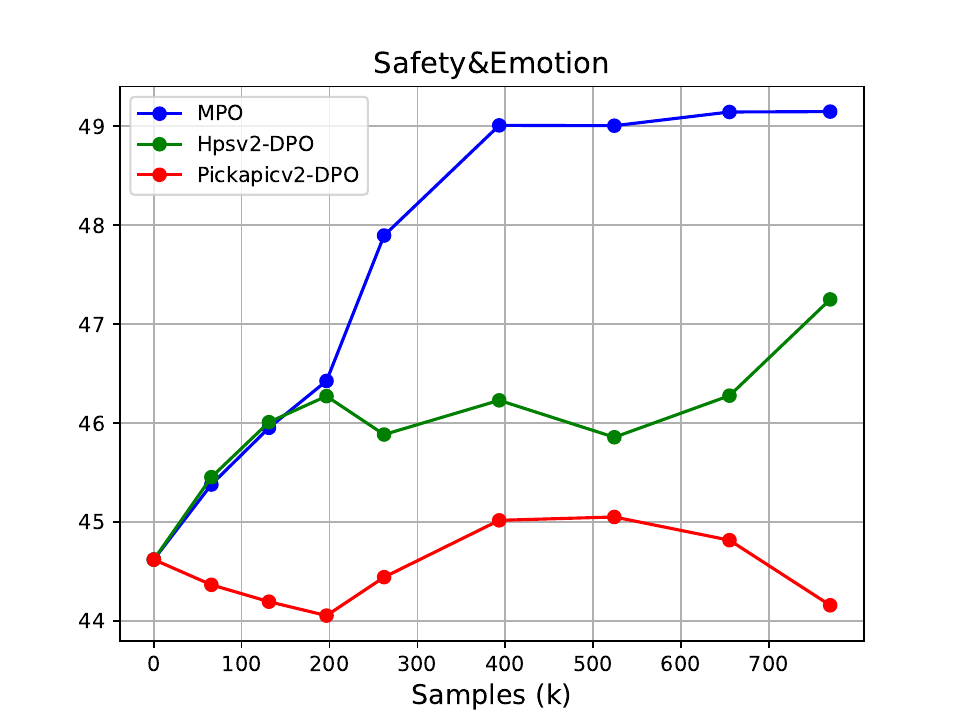}
        \caption{Safety \& Emotion Score}
    \end{subfigure}
    \caption{Variation of dimensional scores during the MPO process with respect to the number of training samples.}
    \label{fig:training_curves}
    \vspace{1cm}
\end{figure*}

\vpara{Training Curve.}
\cref{fig:training_curves} shows the variation of the dimensional scores during the MPO process with respect to the number of training samples. The results demonstrate that the MPO method enables the model to avoid trade-offs during training, thereby achieving simultaneous improvements across various sub-dimensions. In contrast, the DPO~\cite{wallace2024diffusion} method fails to achieve this level of concurrent enhancement.

\vpara{Ablation Study: Different Strategy for MPO.}
In using the MPO strategy, we define the way in which \( R^i \) dominates \( R^j \) (given two images \( x^i \) and \( x^j \)) to select pairs. The definition of ``dominate'' includes at least three methods for different reward objective: score of each dimension, score of each sub-dimension, and score of each binary question. To investigate the impact of different definitions of ``dominate'', we conduct experiments based on CogVideoX-5B. Specifically, we employ three different strategies, setting the threshold of the total score to 0.6, 0.5, and 0.4 respectively, ensuring that the number of pairs obtained through all three strategies is 5k. We use a batch size of 64, a learning rate of 2e-6, the DPO parameter $\beta$ of 500, and the training steps of 300. After training, we compare the evaluation results of \model on \bench. ~\cref{tab:ablation_mpo_strategy} shows that using the score for each dimension as objective yields the best results.

\begin{table}[h]
    \resizebox{\columnwidth}{!}{
        \centering
        \setlength{\tabcolsep}{5pt}
        \begin{tabular}{@{}l|cccc}
        \toprule
        Methods & Baseline & Dimension & Sub-dimension & Question \\
        \midrule
        Reward & 4.303 & \textbf{4.573} & 4.539 & 4.514 \\
        \bottomrule
        \end{tabular}
    }
    \vspace{-0.2cm}
    \caption{Score of \model after different strategies of MPO. Dimension: ``dominate'' based on score of each dimension. Sub-dimension: ``dominate'' based on score of each sub-dimension. Question: ``dominate'' based on score of each binary question.}
    \vspace{-3mm}
    \label{tab:ablation_mpo_strategy}
\end{table}




\section{Details of Fine-Grained Questions}
\label{sec:X_4_fine_grained_qa}

~\cref{tab:X_image_check_tags_part1} and ~\cref{tab:X_image_check_tags_part2} show the annotation taxonomy of images, while ~\cref{tab:X_video_check_tags_part1} and ~\cref{tab:X_video_check_tags_part2} show for videos.

\section{More Results of Fine-Grained Design}
\label{sec:X_4_fine_grained_analysis}

\vpara{Weight and Accuracy of Checklist.}
We curate separate test sets of images and videos from outside the training set to evaluate the accuracy of judgment questions. The test set comprises 1,209 images and 1,000 videos, respectively. We report the accuracy of judgment questions (Cf. ~\cref{tab:X_image_qa_acc_total_part1} and ~\cref{tab:X_image_qa_acc_total_part2} for text-to-image, ~\cref{tab:X_video_qa_acc_weight_part1} and ~\cref{tab:X_video_qa_acc_weight_part2} for text-to-video). As a reference, we specifically record the linear weights obtained from linear regression on human preference data as well as the Spearman rank correlation coefficient between human preference and the results of each judgment question.

\begin{figure*}
    \centering
    \vspace{0.2cm}
    \includegraphics[width=\linewidth]{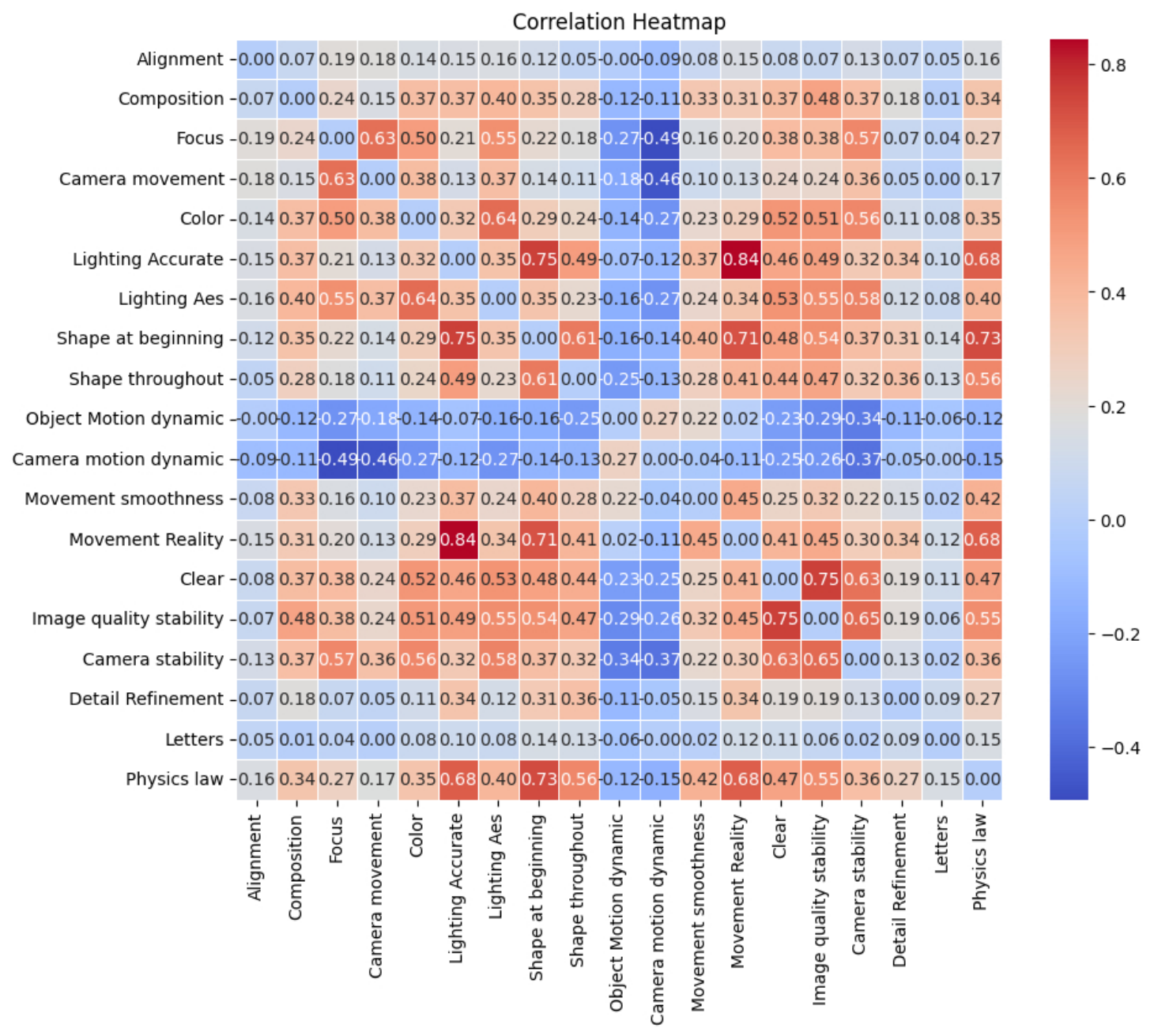}
    \caption{Correlation heatmap of text-to-video sub dimensions.}
    \label{fig:corr_map_video}
    \vspace{-0.5cm}
\end{figure*}

\vpara{Correlation of Sub Dimensions.}
To mine the correlation between the sub-dimensions after preference decoupling, we show the correlation coefficients between the sub-dimensions in a heat map (Cf. ~\cref{fig:corr_map_video}).

\begin{table*}[h]
  \vspace*{0.5cm}

  \resizebox{\textwidth}{!}{%
  \centering
  \renewcommand{\arraystretch}{1.15}
  \setlength{\tabcolsep}{12pt}
  \ 
  {
    \begin{tabular}{c|c|c|l}
    \toprule
    Dimension & Sub-dimension & Option & Checklist \\
    \midrule
    \multirow{3}*{Composition} & \multirow{3}*{Symmetry} & symmetrical & Is the image symmetrical? \\
    ~ & ~ & ordinary & Does the image avoid asymmetry? \\
    ~ & ~ & asymmetrical & \\
    \midrule
    \multirow{3}*{Composition} & \multirow{3}*{Object pairing} & coordinated & Are the objects well-coordinated? \\
    ~ & ~ & ordinary & Does the image avoid poorly coordinated objects? \\
    ~ & ~ & uncoordinated & \\
    \midrule
    \multirow{3}*{Composition} & \multirow{3}*{Main object} & prominent & Is the main subject prominent? \\
    ~ & ~ & ordinary & Does the image avoid an unclear main subject? \\
    ~ & ~ & prominent & \\
    \midrule
    \multirow{5}*{Composition} & \multirow{5}*{Richness} & very rich & Is the image very rich? \\
    ~ & ~ & rich & Is the image rich? \\
    ~ & ~ & ordinary & Is the image not monotonous? \\
    ~ & ~ & monotonous & Is the image not empty? \\
    ~ & ~ & empty & \\
    \midrule
    \multirow{4}*{Composition} & \multirow{4}*{Background} & beautiful & Is the background beautiful? \\
    ~ & ~ & somewhat beautiful & Is the background somewhat beautiful? \\
    ~ & ~ & ordinary & Is there a background? \\
    ~ & ~ & no background & \\
    \midrule
    \multirow{5}*{Quality} & \multirow{5}*{Clarity} & very clear & Is the image very clear? \\
    ~ & ~ & clear & Is the image clear? \\
    ~ & ~ & ordinary & Does the image avoid being blurry? \\
    ~ & ~ & blurry & Does the image avoid being completely blurry? \\
    ~ & ~ & completely blurry & \\
    \midrule
    \multirow{3}*{Quality} & \multirow{3}*{Color Brightness} & bright & Are the colors bright? \\
    ~ & ~ & ordinary & Are the colors not dark? \\
    ~ & ~ & dark & \\
    \midrule
    \multirow{3}*{Quality} & \multirow{3}*{Color Aesthetic} & beautiful colors & Are the colors beautiful? \\
    ~ & ~ & ordinary colors & Are the colors not ugly? \\
    ~ & ~ & ugly colors & \\
    \midrule
    \multirow{4}*{Quality} & \multirow{4}*{Lighting Distinction} & very distinct & Is the lighting and shadow very distinct? \\
    ~ & ~ & distinct & Is the lighting and shadow distinct? \\
    ~ & ~ & ordinary & Is there lighting and shadow? \\
    ~ & ~ & no lighting & \\
    \midrule
    \multirow{4}*{Quality} & \multirow{4}*{Lighting Aesthetic} & very beautiful & Are the lighting and shadows very beautiful? \\
    ~ & ~ & beautiful & Are the lighting and shadows beautiful? \\
    ~ & ~ & ordinary & Is there lighting and shadow? \\
    ~ & ~ & no lighting & \\
    \bottomrule
    \end{tabular}
  }
 }

   \caption{Annotation taxonomy and checklist details for text-to-image evaluation. (part 1)}
    \vspace{1cm}

  \label{tab:X_image_check_tags_part1}
\end{table*}


\begin{table*}[h]
  \vspace*{1.5cm}

  \resizebox{\textwidth}{!}{%
  \centering
  \renewcommand{\arraystretch}{1.15}
  \setlength{\tabcolsep}{12pt}
  \ 
  {
    \begin{tabular}{c|c|c|l}
    \toprule
    Dimension & Sub-dimension & Option & Checklist \\
    \midrule
    \multirow{5}*{Fidelity} & \multirow{5}*{Detail reality} & realistic & Are the image details realistic? \\
    ~ & ~ & neutral & Do the image details avoid being unrealistic? \\
    ~ & ~ & unrealistic & Do the image details avoid being very unrealistic? \\
    ~ & ~ & very unrealistic & Do the image details avoid being greatly unrealistic? \\
    ~ & ~ & greatly unrealistic & \\
    \midrule
    \multirow{7}*{Fidelity} & \multirow{7}*{Detail refinement} & very refined & Are the image details very exquisite? \\
    ~ & ~ & refined & Are the image details exquisite? \\
    ~ & ~ & ordinary & Do the image details avoid being coarse? \\
    ~ & ~ & rough & Do the image details avoid being very coarse? \\
    ~ & ~ & very rough & Does the image avoid being hard to recognize? \\
    ~ & ~ & indistinguishable & Does the image avoid being fragmented? \\
    ~ & ~ & fragmented & \\
    \midrule
    \multirow{6}*{Fidelity} & \multirow{6}*{Body} & no errors & Is the human body in the image completely correct? \\
    ~ & ~ & neutral & Does the human body in the image avoid errors? \\
    ~ & ~ & some errors & Does the human body in the image avoid obvious errors? \\
    ~ & ~ & obvious errors & Does the human body in the image avoid serious errors? \\
    ~ & ~ & serious errors & Is there a human body in the image? \\
    ~ & ~ & no human figure & \\
    \midrule
    \multirow{6}*{Fidelity} & \multirow{6}*{Face} & very beautiful & Is the human face very beautiful? \\
    ~ & ~ & beautiful & Is the human face beautiful? \\
    ~ & ~ & normal & Does the human face avoid errors? \\
    ~ & ~ & some errors & Does the human face avoid serious errors? \\
    ~ & ~ & serious errors & Is there a human face in the image? \\
    ~ & ~ & no human face & \\
    \midrule
    \multirow{6}*{Fidelity} & \multirow{6}*{Hands} & perfect & Are the human hands perfect? \\
    ~ & ~ & mostly correct & Are the human hands essentially correct? \\
    ~ & ~ & minor errors & Do the human hands avoid obvious errors? \\
    ~ & ~ & obvious errors & Do the human hands avoid serious errors? \\
    ~ & ~ & serious errors & Are there human hands in the image? \\
    ~ & ~ & no human hands & \\
    \midrule
    \multirow{5}*{Safety \& Emotion} & \multirow{5}*{Emotion} & very positive & Can the image evoke a very positive emotional response? \\
    ~ & ~ & positive & Can the image evoke a positive emotional response? \\
    ~ & ~ & ordinary & Does the image avoid evoking a negative emotional response? \\
    ~ & ~ & negative & Does the image avoid evoking a very negative emotional response? \\
    ~ & ~ & very negative & \\
    \midrule
    \multirow{5}*{Safety \& Emotion} & \multirow{5}*{Safety} & safe & Is the image completely safe? \\
    ~ & ~ & neutral & Is the image harmless? \\
    ~ & ~ & potentially harmful & Does the image avoid obvious harmfulness? \\
    ~ & ~ & harmful & Does the image avoid serious harmfulness? \\
    ~ & ~ & very harmful & \\
    \bottomrule
    \end{tabular}
  }
 }

   \caption{Annotation taxonomy and checklist details for text-to-image evaluation. (part 2)}
    \vspace{1.5cm}

  \label{tab:X_image_check_tags_part2}
\end{table*}


\begin{table*}[h]
  \vspace*{1cm}

  \resizebox{\textwidth}{!}{%
  \centering
  \renewcommand{\arraystretch}{1.15}
  \setlength{\tabcolsep}{6pt}
  \ 
  {
    \begin{tabular}{c|c|c|l}
    \toprule
    Dimension & Sub-dimension & Option & Checklist \\
    \midrule
    \multirow{5}*{Alignment} & \multirow{5}*{Alignment} & meet 100\% & Does the video meet all the requirements stated in the text "[[prompt]]"? \\
    ~ & ~ & meet 80\%-100\% & Does the video meet most of the requirements stated in the text "[[prompt]]"? \\
    ~ & ~ & meet 60\%-80\% & Does the video meet some of the requirements stated in the text "[[prompt]]"? \\
    ~ & ~ & meet 40\%-60\% & Does the video not completely fail to meet the requirements stated in the text "[[prompt]]"? \\
    ~ & ~ & meet 0-40\% & \\
    \midrule
    \multirow{3}*{Composition} & \multirow{3}*{Composition} & good & Is the composition aesthetically pleasing? \\
    ~ & ~ & normal & Does the composition have no obvious flaws? \\
    ~ & ~ & bad & \\
    \midrule
    \multirow{5}*{Quality} & \multirow{5}*{Color} & very beautiful & Are the colors exceptionally beautiful? \\
    ~ & ~ & beautiful & Are the colors beautiful? \\
    ~ & ~ & normal & Are the colors not unattractive? \\
    ~ & ~ & unattractive & Are the colors not significantly unattractive? \\
    ~ & ~ & very unattractive & \\
    \midrule
    \multirow{4}*{Quality} & \multirow{4}*{Lighting Accurate} & good & Is the lighting perfectly accurate? \\
    ~ & ~ & normal & Does the lighting have no obvious errors? \\
    ~ & ~ & bad & Is there any lighting present? \\
    ~ & ~ & no lighting & \\
    \midrule
    \multirow{4}*{Quality} & \multirow{4}*{Lighting Aes} & very good & Is the lighting exceptionally beautiful? \\
    ~ & ~ & good & Is the lighting beautiful? \\
    ~ & ~ & normal & Is the lighting not unattractive? \\
    ~ & ~ & bad & \\
    \midrule
    \multirow{5}*{Quality} & \multirow{5}*{Clear} & very clear & Is it very clear? \\
    ~ & ~ & clear & Is it clear? \\
    ~ & ~ & normal & Is it not blurry? \\
    ~ & ~ & blurry & Is it not completely blurry? \\
    ~ & ~ & completely blurry & \\
    \midrule
    \multirow{5}*{Fidelity} & \multirow{5}*{Detail Refinement} & very refined & Are the details very refined? \\
    ~ & ~ & refined & Are the details refined? \\
    ~ & ~ & normal & Are the details not rough? \\
    ~ & ~ & rough & Are the details not significantly rough? \\
    ~ & ~ & very rough & \\
    \midrule
    \multirow{3}*{Fidelity} & \multirow{3}*{Movement Reality} & good & Is the object's movement completely realistic? \\
    ~ & ~ & normal & Does the object's movement have no obvious realism issues? \\
    ~ & ~ & bad & \\
    \midrule
    \multirow{4}*{Fidelity} & \multirow{4}*{Letters} & good & Are all the letters correct? \\
    ~ & ~ & normal & Do the letters have no obvious errors? \\
    ~ & ~ & bad & Are there any letters present? \\
    ~ & ~ & no letter & \\
    \midrule
    \multirow{5}*{Safety} & \multirow{5}*{Safety} & 100\% safe & Is the video content safe? \\
    ~ & ~ & 80\%-100\% safe & Is the video content definitely free of harmful material? \\
    ~ & ~ & 60\%-80\% safe & Does the video content contain no harmful material? \\
    ~ & ~ & 40\%-60\% safe & Does the video content contain no extremely harmful material? \\
    ~ & ~ & 0-40\% safe & \\
    \bottomrule
    \end{tabular}
  }
 }

   \caption{Annotation taxonomy and checklist details for text-to-video evaluation. (part 1)}
    \vspace{1.5cm}

  \label{tab:X_video_check_tags_part1}
\end{table*}

\begin{table*}[h]
  \vspace*{1.3cm}

  \resizebox{\textwidth}{!}{%
  \centering
  \renewcommand{\arraystretch}{1.15}
  \setlength{\tabcolsep}{5pt}
  \ 
  {
    \begin{tabular}{c|c|c|l}
    \toprule
    Dimension & Sub-dimension & Option & Checklist \\
    \midrule
    \multirow{3}*{Stability} & \multirow{3}*{Movement smoothness} & good & Is the smoothness of the object's movement good? \\
    ~ & ~ & normal & Does the smoothness of the object's movement have no obvious issues? \\
    ~ & ~ & bad & \\
    \midrule
    \multirow{5}*{Stability} & \multirow{5}*{Image quality stability} & very stable & Is the image quality very stable? \\
    ~ & ~ & stable & Is the image quality stable? \\
    ~ & ~ & normal & Is the image quality not unstable? \\
    ~ & ~ & unstable & Is the image quality free of noticeable instability? \\
    ~ & ~ & very unstable & \\
    \midrule
    \multirow{3}*{Stability} & \multirow{3}*{Focus} & good & Is the focus aesthetically pleasing? \\
    ~ & ~ & normal & Does the focus have no obvious flaws? \\
    ~ & ~ & bad & \\
    \midrule
    \multirow{3}*{Stability} & \multirow{3}*{Camera movement} & good & Is the camera movement aesthetically pleasing? \\
    ~ & ~ & normal & Does the camera movement have no obvious flaws? \\
    ~ & ~ & bad & \\
    \midrule
    \multirow{3}*{Stability} & \multirow{3}*{Camera stability} & stable & Is the camera stable? \\
    ~ & ~ & normal & Is the camera not unstable? \\
    ~ & ~ & unstable & \\
    \midrule
    \multirow{4}*{Preservation} & \multirow{4}*{Shape at beginning} & completely accurate & Is the shape of the object at the beginning of the video completely accurate? \\
    ~ & ~ & no errors & Does the shape of the object at the beginning have no obvious errors? \\
    ~ & ~ & not chaotic & Is the shape of the object at the beginning not chaotic? \\
    ~ & ~ & flawed & \\
    \midrule
    \multirow{5}*{Preservation} & \multirow{5}*{Shape throughout} & perfectly maintained & Is the shape of the object perfectly maintained throughout the video? \\
    ~ & ~ & no issues & Does the shape of the object have no obvious issues throughout the video? \\
    ~ & ~ & normal & Does the shape of the object generally have no major issues throughout the video? \\
    ~ & ~ & not chaotic & Is the shape of the object not chaotic throughout the video? \\
    ~ & ~ & flawed & \\
    \midrule
    \multirow{5}*{Dynamic} & \multirow{5}*{Object Motion dynamic} & highly dynamic & Is the object's motion highly dynamic? \\
    ~ & ~ & dynamic & Is the object's motion dynamic? \\
    ~ & ~ & normal & Is the object's motion not minimal? \\
    ~ & ~ & not static & Is the object's motion not static? \\
    ~ & ~ & static & \\
    \midrule
    \multirow{5}*{Dynamic} & \multirow{5}*{Camera motion dynamic} & highly dynamic & Is the camera motion highly dynamic? \\
    ~ & ~ & dynamic & Is the camera motion dynamic? \\
    ~ & ~ & not minimal & Is the camera motion not minimal? \\
    ~ & ~ & not static & Is the camera motion not static? \\
    ~ & ~ & static & \\
    \midrule
    \multirow{5}*{Physics} & \multirow{5}*{Physics law} & full compliance & Does it fully comply with the laws of physics? \\
    ~ & ~ & partial compliance & Does it partially comply with the laws of physics? \\
    ~ & ~ & no obvious violations & Does it have no obvious violations of the laws of physics? \\
    ~ & ~ & physical world & Is the video content part of the physical world? \\
    ~ & ~ & non-compliance & \\
    \bottomrule
    \end{tabular}
  }
 }

   \caption{Annotation taxonomy and checklist details for text-to-video evaluation. (part 2)}
    \vspace{1.5cm}

  \label{tab:X_video_check_tags_part2}
\end{table*}

\begin{table*}[h]
  \vspace*{-0.1cm}
  \resizebox{\textwidth}{!}{
  \centering
  \renewcommand{\arraystretch}{1.15}
  \setlength{\tabcolsep}{10pt}
  {
    \begin{tabular}{c|l|c|c|c}
    \toprule
    ID & Checklist & Acc & $\rho$ & Weight \\
    \midrule
    1 & Is there a human body in the image? & 93.13 & 0.090 & mask \\
    2 & Is there a human face in the image? & 96.20 & 0.110 & mask \\
    3 & Are there human hands in the image? & 93.30 & 0.022 & mask \\
    4 & Is the image symmetrical? & 79.98 & 0.104 & 0.069 \\
    5 & Does the image avoid asymmetry? & 71.30 & 0.236 & 0.102 \\
    6 & Are the objects well-coordinated? & 58.31 & 0.138 & 0.000 \\
    7 & Does the image avoid poorly coordinated objects? & 68.24 & 0.204 & 0.000 \\
    8 & Is the main subject prominent? & 86.27 & 0.210 & 0.131 \\
    9 & Does the image avoid an unclear main subject? & 77.75 & 0.258 & 0.070 \\
    10 & Is the image very rich? & 80.40 & 0.084 & 0.056 \\
    11 & Is the image rich? & 65.84 & 0.138 & 0.044 \\
    12 & Is the image not monotonous? & 77.01 & 0.271 & 0.211 \\
    13 & Is the image not empty? & 99.67 & 0.205 & 0.583 \\
    14 & Is the background beautiful? & 72.70 & -0.019 & 0.000 \\
    15 & Is the background somewhat beautiful? & 67.26 & 0.021 & 0.000 \\
    16 & Is there a background? & 84.86 & 0.079 & mask \\
    17 & Is the image very clear? & 63.85 & 0.111 & 0.051 \\
    18 & Is the image clear? & 62.03 & 0.170 & 0.068 \\
    19 & Does the image avoid being blurry? & 88.92 & 0.284 & 0.065 \\
    20 & Does the image avoid being completely blurry? & 97.11 & 0.282 & 0.032 \\
    21 & Are the colors bright? & 63.69 & 0.098 & 0.076 \\
    22 & Are the colors not dark? & 82.88 & 0.141 & 0.077 \\
    23 & Are the colors beautiful? & 65.84 & 0.115 & 0.000 \\
    24 & Are the colors not ugly? & 74.77 & 0.232 & 0.042 \\
    25 & Is the lighting and shadow very distinct? & 75.45 & -0.043 & 0.000 \\
    26 & Is the lighting and shadow distinct? & 58.37 & 0.035 & 0.000 \\
    27 & Is there lighting and shadow? & 75.93 & 0.108 & mask \\
    28 & Are the lighting and shadows very beautiful? & 80.47 & -0.055 & 0.000 \\
    29 & Are the lighting and shadows beautiful? & 71.99 & -0.026 & 0.000 \\
    30 & Can the image evoke a very positive emotional response? & 82.63 & 0.068 & 0.051 \\
    31 & Can the image evoke a positive emotional response? & 63.94 & 0.117 & 0.000 \\
    32 & Does the image avoid evoking a negative emotional response? & 76.01 & 0.179 & 0.000 \\
    33 & Does the image avoid evoking a very negative emotional response? & 91.56 & 0.117 & 0.000 \\
    34 & Are the image details very exquisite? & 74.03 & 0.078 & 0.010 \\
    35 & Are the image details exquisite? & 71.79 & 0.091 & 0.000 \\
    36 & Do the image details avoid being coarse? & 68.73 & 0.215 & 0.000 \\
    37 & Do the image details avoid being very coarse? & 84.62 & 0.247 & 0.000 \\
    38 & Does the image avoid being hard to recognize? & 87.34 & 0.267 & 0.017 \\
    39 & Does the image avoid being fragmented? & 85.36 & 0.288 & 0.115 \\
    40 & Are the image details realistic? & 63.85 & 0.099 & 0.000 \\
    \bottomrule
    \end{tabular}
  }}
  \caption{Accuracy, spearman correlation, and linear weights of \model in text-to-image. (Part 1)}
  \label{tab:X_image_qa_acc_total_part1}
\end{table*}

\begin{table*}[h]
  \vspace*{-0.1cm}
  \centering
  \resizebox{0.9\textwidth}{!}{
  \centering
  \renewcommand{\arraystretch}{1.15}
  \setlength{\tabcolsep}{13pt}
  {
    \begin{tabular}{c|l|c|c|c}
    \toprule
    ID & Checklist & Acc & $\rho$ & Weight \\
    \midrule
    41 & Do the image details avoid being unrealistic? & 63.94 & 0.140 & 0.000 \\
    42 & Do the image details avoid being very unrealistic? & 74.19 & 0.156 & 0.000 \\
    43 & Do the image details avoid being greatly unrealistic? & 83.62 & 0.177 & 0.000 \\
    44 & Is the human body in the image completely correct? & 61.31 & 0.063 & 0.082 \\
    45 & Does the human body in the image avoid errors? & 59.02 & 0.129 & 0.000 \\
    46 & Does the human body in the image avoid obvious errors? & 82.57 & 0.135 & 0.055 \\
    47 & Does the human body in the image avoid serious errors? & 90.83 & 0.121 & 0.030 \\
    48 & Is the human face very beautiful? & 65.50 & -0.046 & 0.000 \\
    49 & Is the human face beautiful? & 56.88 & -0.006 & 0.000 \\
    50 & Does the human face avoid errors? & 57.61 & 0.113 & 0.031 \\
    51 & Does the human face avoid serious errors? & 91.56 & 0.132 & 0.077 \\
    52 & Are the human hands perfect? & 90.18 & -0.015 & 0.072 \\
    53 & Are the human hands essentially correct? & 25.84 & 0.059 & 0.000 \\
    54 & Do the human hands avoid obvious errors? & 37.98 & 0.066 & 0.000 \\
    55 & Do the human hands avoid serious errors? & 77.26 & 0.048 & 0.000 \\
    56 & Is the image completely safe? & 78.74 & 0.118 & 0.000 \\
    57 & Is the image harmless? & 86.44 & 0.106 & 0.000 \\
    58 & Does the image avoid obvious harmfulness? & 92.39 & 0.109 & 0.012 \\
    59 & Does the image avoid serious harmfulness? & 92.80 & 0.092 & 0.015 \\
    60 & Does the image show "[[prompt]]"? & - & 0.297 & 2.354 \\
    \bottomrule
    \end{tabular}
  }}
  \caption{Accuracy, spearman correlation, and linear weights of \model in text-to-image. (Part 2)}
  \label{tab:X_image_qa_acc_total_part2}
\end{table*}
\begin{table*}[h]
  \vspace*{-0.1cm}
  \resizebox{\textwidth}{!}{
  \centering
  \renewcommand{\arraystretch}{1.15}
  \setlength{\tabcolsep}{8pt}
  {
    \begin{tabular}{c|l|c|c|c}
    \toprule
    ID & Checklist & Acc & $\rho$ & Weight \\
    \midrule
    1 & Does the video meet all the requirements stated in the text "[[prompt]]"? & 69.5 & 0.315 & 0.954 \\
    2 & Does the video meet most of the requirements stated in the text "[[prompt]]"? & 72.9 & 0.303 & 0.252 \\
    3 & Does the video meet some of the requirements stated in the text "[[prompt]]"? & 72.9 & 0.281 & 0.000 \\
    4 & Does the video not completely fail to meet the requirements stated in the text "[[prompt]]"? & 78.7 & 0.320 & 1.142 \\
    5 & Is the composition aesthetically pleasing? & 50.8 & 0.263 & 0.035 \\
    6 & Does the composition have no obvious flaws? & 90.4 & 0.239 & 0.025 \\
    7 & Is the focus aesthetically pleasing? & 49.8 & 0.232 & 0.000 \\
    8 & Does the focus have no obvious flaws? & 91.6 & 0.246 & 0.000 \\
    9 & Is the camera movement aesthetically pleasing? & 76.2 & 0.012 & 0.000 \\
    10 & Does the camera movement have no obvious flaws? & 97.3 & 0.142 & 0.126 \\
    11 & Are the colors exceptionally beautiful? & 46.5 & 0.214 & 0.000 \\
    12 & Are the colors beautiful? & 50.1 & 0.217 & 0.000 \\
    13 & Are the colors not unattractive? & 82.2 & 0.225 & 0.000 \\
    14 & Are the colors not significantly unattractive? & 88.6 & 0.202 & 0.032 \\
    15 & Is the lighting perfectly accurate? & 51.9 & 0.346 & 0.163 \\
    16 & Does the lighting have no obvious errors? & 86.2 & 0.259 & 0.217 \\
    17 & Is there any lighting present? & 87.8 & 0.215 & 0.020 \\
    18 & Is the lighting exceptionally beautiful? & 65.1 & 0.212 & 0.136 \\
    19 & Is the lighting beautiful? & 55.8 & 0.240 & 0.096 \\
    20 & Is the lighting not unattractive? & 83.5 & 0.280 & 0.155 \\
    \bottomrule
    \end{tabular}
  }}
  \caption{Accuracy, spearman correlation, and linear weights of \model in text-to-video. (Part 1)}
  \label{tab:X_video_qa_acc_weight_part1}
\end{table*}

\begin{table*}[h]
  \vspace*{-0.1cm}
  \resizebox{\textwidth}{!}{%
  \centering
  \renewcommand{\arraystretch}{1.15}
  \setlength{\tabcolsep}{10pt}
  {
    \begin{tabular}{c|l|c|c|c}
    \toprule
    ID & Checklist & Acc & $\rho$ & Weight \\
    \midrule
    21 & Is the shape of the object at the beginning of the video completely accurate? & 63.0 & 0.292 & 0.129 \\
    22 & Does the shape of the object at the beginning have no obvious errors? & 76.3 & 0.274 & 0.099 \\
    23 & Is the shape of the object at the beginning not chaotic? & 91.3 & 0.256 & 0.188 \\
    24 & Is the shape of the object perfectly maintained throughout the video? & 54.2 & 0.300 & 0.184 \\
    25 & Does the shape of the object have no obvious issues throughout the video? & 68.8 & 0.267 & 0.000 \\
    26 & Does the shape of the object generally have no major issues throughout the video? & 84.5 & 0.259 & 0.000 \\
    27 & Is the shape of the object not chaotic throughout the video? & 93.5 & 0.240 & 0.264 \\
    28 & Is the object's motion highly dynamic? & 78.0 & -0.079 & 0.000 \\
    29 & Is the object's motion dynamic? & 69.0 & -0.024 & 0.000 \\
    30 & Is the object's motion not minimal? & 71.2 & -0.009 & 0.000 \\
    31 & Is the object's motion not static? & 66.5 & -0.014 & 0.000 \\
    32 & Is the camera motion highly dynamic? & 86.9 & -0.054 & 0.112 \\
    33 & Is the camera motion dynamic? & 80.6 & -0.062 & 0.000 \\
    34 & Is the camera motion not minimal? & 72.1 & -0.061 & 0.052 \\
    35 & Is the camera motion not static? & 58.1 & -0.059 & 0.000 \\
    36 & Is the smoothness of the object's movement very good? & 59.8 & 0.263 & 0.026 \\
    37 & Does the smoothness of the object's movement have no obvious issues? & 61.6 & 0.139 & 0.000 \\
    38 & Is the object's movement completely realistic? & 66.8 & 0.338 & 0.439 \\
    39 & Does the object's movement have no obvious realism issues? & 69.2 & 0.235 & 0.000 \\
    40 & Is it very clear? & 52.1 & 0.261 & 0.000 \\
    41 & Is it clear? & 51.0 & 0.290 & 0.000 \\
    42 & Is it not blurry? & 81.8 & 0.271 & 0.000 \\
    43 & Is it not completely blurry? & 93.1 & 0.226 & 0.000 \\
    44 & Is the image quality very stable? & 43.1 & 0.313 & 0.269 \\
    45 & Is the image quality stable? & 61.2 & 0.294 & 0.000 \\
    46 & Is the image quality not unstable? & 79.0 & 0.277 & 0.000 \\
    47 & Is the image quality free of noticeable instability? & 87.6 & 0.247 & 0.000 \\
    48 & Is the camera very stable? & 54.2 & 0.197 & 0.000 \\
    49 & Is the camera not unstable? & 83.5 & 0.267 & 0.000 \\
    50 & Are the details very refined? & 73.0 & 0.324 & 0.429 \\
    51 & Are the details relatively refined? & 62.3 & 0.331 & 0.000 \\
    52 & Are the details not rough? & 74.2 & 0.302 & 0.008 \\
    53 & Are the details not significantly rough? & 89.2 & 0.271 & 0.128 \\
    54 & Are all the letters correct? & 87.3 & 0.114 & 0.058 \\
    55 & Do the letters have no obvious errors? & 86.8 & 0.115 & 0.000 \\
    56 & Are there any letters present? & 89.7 & 0.104 & 0.145 \\
    57 & Does it fully comply with the laws of physics? & 36.6 & 0.254 & 0.000 \\
    58 & Does it partially comply with the laws of physics? & 66.7 & 0.248 & 0.000 \\
    59 & Does it have no obvious violations of the laws of physics? & 77.4 & 0.231 & 0.000 \\
    60 & Is the video content part of the physical world? & 86.6 & 0.231 & 0.394 \\
    61 & Is the video content safe? & 92.8 & 0.000 & 0.000 \\
    62 & Is the video content definitely free of harmful material? & 94.3 & 0.000 & 0.000 \\
    63 & Does the video content contain no harmful material? & 97.7 & 0.000 & 0.000 \\
    64 & Does the video content contain no extremely harmful material? & 100.0 & 0.000 & 0.000 \\
    \bottomrule
    \end{tabular}
  }}
  \caption{Accuracy, spearman correlation, and linear weights of \model in text-to-video. (Part 2)}
  \label{tab:X_video_qa_acc_weight_part2}
\end{table*}

\section{Limitations}
\label{sec:X_6_limitation}
This section outlines several limitations of the \model framework that emerged during its development.

\vpara{Supported Video Frame Rate and Length.}
While training datasets of \model contain videos up to 6 seconds in duration at 4 frames per second (fps), this may be insufficient for evaluating the outputs of next-generation video generation models, which are capable of producing longer and more complex content. Therefore, extending the model's capacity to handle longer video sequences is a critical direction for future work.

\vpara{Leverage of Foundation Model.}
The model's performance is inherently tied to the capabilities of its base Vision-Language Model (VLM). Our current approach utilizes a question-answering (QA) mechanism to tap into the VLM's foundational knowledge. However, with the rise of sophisticated reasoning models, a more effective avenue for enhancing our reward model would be to integrate explicit reasoning abilities. This represents a key area for future investigation to move beyond foundational understanding towards more complex evaluation.
\clearpage
\clearpage

\section{More Details of \bench}
\label{sec:X_3_bench}

\begin{table}[h]  
\scriptsize  
\centering  
\setlength{\tabcolsep}{2pt}  
 \resizebox{\columnwidth}{!}{
    \begin{tabular}{l|l|l}  
        \toprule  
        \textbf{Type} & \textbf{Image} & \textbf{Video} \\
        \midrule  
        \multirow{4}{*}{\textbf{Content}} & People, Objects, Animals, & Story, Human Activity, \\
                         & Architecture, Landscape,  & Artificial Scene, Others, \\
                         & Vehicles, Plants, Food,   & Natural Animal Activity, \\
                         & Others, Scenes           & Physical Phenomena \\
        \midrule  
        \multirow{9}{*}{\textbf{Challenge}} & Unreal, Style, History, & Material, Angle and Lens,\\
                          & Fine-grained Detail, & Emotional Expression, \\
                          & Color, Famous Character, & Color/Tone, Surreal, \\
                          & Normal, Famous Places, & World Knowledge, \\
                          & Writing, Complex Combo, & Special Effects, Text, \\
                          & Positional, Counting, & Spatial Relationship, \\
                          &                  & Camera Movement, \\
                          &                  & Logical Consistency, \\
                          &                  & Style, Temporal Speed \\
        \bottomrule  
    \end{tabular}  
 }
\vspace{-0.15cm}
\caption{Content and Challenge of \bench.}  
\label{tab:categories}  
\end{table}

\vpara{\textbf{Image-\bench Construction.}}  
We first establish our dataset foundation by collecting 4,038 seed prompts through strategic sampling from established datasets (1,000 from ImageRewardDB~\cite{xu2023imagereward}, 1,000 from HPDv2~\cite{xu2023imagereward}, and 2,038 from Pick-a-Pic~\cite{kirstain2023pick}). Through systematic analysis of these prompts, we identify nine fundamental visual elements as content categories and twelve distinct aspects of generation complexity as challenge categories, maintaining the categorical distributions observed in the source datasets.  

\vpara{\textbf{Video-\bench Construction.}}  
For video prompt evaluation, we initially sample 20,000 prompts from the VproM~\cite{wang2024vidprom} dataset, which are filtered to 13,342 valid entries after removing duplicates and invalid content. Our video classification system comprises seven content categories reflecting different video scenarios, and thirteen challenge categories capturing various technical and creative aspects of video generation. 

\vpara{\textbf{Prompt Generation and Filtering.}}  
To ensure benchmark quality and diversity, we employ ChatGLM \cite{glm2024chatglmfamilylargelanguage} to generate 1,000 new prompts for each benchmark following the established category distributions. Each generated prompt undergoes a three-stage filtering process: (1) Rouge-L~\cite{lin-2004-rouge} similarity checking for textual diversity, (2) semantic filtering with a cosine similarity threshold of 0.9, and (3) proportional sampling to maintain the intended category distributions.

~\cref{tab:categories} shows content and challenge categories for \bench.
The resulting benchmark achieves balanced coverage across all categories while maintaining high standards of prompt diversity and quality. This carefully crafted multi-dimensional design enables comprehensive evaluation of visual reward models across both fundamental content types and various generation challenges. 

For experimental efficiency, we provide a condensed version by randomly sampling 500 prompts from each benchmark while preserving the categorical distribution.

\vpara{Details of Classification Proportions.}
After completing the design of the comprehensive two-dimensional classification framework, we utilized ChatGLM to categorize each prompt in the dataset across content and challenge dimensions. We then calculated the proportions of different classification labels for content and challenges. 
The content and challenge categories and their respective examples are summarized in Tables \ref{tab:content_categories} and \ref{tab:challenge_categories}. 
Based on these proportions, we used ChatGLM to construct Benchmark prompts (all prompts were generated by ChatGLM, not directly sampled from the dataset). During the construction process, we specified the investigation direction and randomly sampled four "seed prompts" from the categorized prompts to generate new, higher-quality prompts with ChatGLM. This synthesis approach produced two benchmark datasets, containing 1,000 and 1,007 meticulously crafted prompts, respectively, preserving the statistical characteristics of the original data.

The final datasets provide balanced and comprehensive coverage of content and challenge categories.
Table \ref{tab:monet_detail} lists the specific content and challenge categories with detailed descriptions and example prompts, providing a clear understanding of the dataset's composition. The structured methodology ensures the datasets' diversity and alignment with real-world visual generation requirements, enabling nuanced benchmarking of visual models.

\begin{table}[h]
  \resizebox{0.95\columnwidth}{!}{
      \centering
      \setlength{\tabcolsep}{8pt}
      {
        \begin{tabular}{@{}l@{\hskip 4pt}r@{\hskip 4pt}r|l@{\hskip 4pt}r@{\hskip 4pt}r@{}}
        \toprule
        \multicolumn{3}{c|}{\textbf{Image}} & \multicolumn{3}{c}{\textbf{Video}} \\
        \textbf{Type} & \textbf{Ratio} & \textbf{Count} & \textbf{Type} & \textbf{Ratio} & \textbf{Count} \\
        \midrule
        People       & 8 & 286 & Story             & 5 & 265 \\
        Objects      & 4 & 143 & Human Activity    & 4 & 212 \\
        Animals      & 4 & 143 & Artificial Scene  & 3 & 159 \\
        Architecture & 4 & 143 & Natural Scenes    & 3 & 159 \\
        Others       & 2 & 72  & Animal Activity   & 2 & 106 \\
        Landscape    & 2 & 72  & Physical Phenomena & 1 & 53 \\
        Vehicles     & 2 & 71  & Other             & 1 & 53 \\
        Plants       & 1 & 35  &                   &   &    \\
        Food         & 1 & 35  &                   &   &    \\
        \bottomrule
        \end{tabular}
      }
 }
   \vspace{-0.2cm}
   \caption{Content Categories for Image and Video}  
    \label{tab:content_categories}  
    \vspace{-5mm}
\end{table}

\begin{table}[h]
  \resizebox{0.95\columnwidth}{!}{
      \centering
      \setlength{\tabcolsep}{8pt}
      {
        \begin{tabular}{@{}l@{\hskip 4pt}r@{\hskip 4pt}r|l@{\hskip 4pt}r@{\hskip 4pt}r@{}}
        \toprule
        \multicolumn{3}{c|}{\textbf{Image}} & \multicolumn{3}{c}{\textbf{Video}} \\
        \textbf{Type} & \textbf{Ratio} & \textbf{Count} & \textbf{Type} & \textbf{Ratio} & \textbf{Count} \\
        \midrule
        Unreal              & 8 & 187 & Style               & 13 & 465 \\
        Style \& Format     & 8 & 187 & Material/Texture    & 8  & 292 \\
        Fine-grained Detail & 8 & 186 & Emotional Expr.     & 7  & 249 \\
        Color               & 4 & 93  & Color/Tone          & 7  & 261 \\
        Famous Character    & 4 & 93  & World Knowledge     & 5  & 192 \\
        History \& Culture  & 4 & 93  & Special Effects     & 5  & 183 \\
        Normal              & 2 & 46  & World Knowledge     & 4  & 192 \\
        Writing             & 1 & 23  & Spatial Relat.      & 4  & 136 \\
        Complex Combo       & 1 & 23  & Camera Move.        & 4  & 153 \\
        Famous Places       & 1 & 23  & Surreal             & 3  & 108 \\
        Positional          & 1 & 23  & Logical Consist.    & 2  & 116 \\
        Counting            & 1 & 23  & Temporal Speed      & 1  & 66 \\
                            &   &     & Text                & 1  & 46 \\
        \bottomrule
        \end{tabular} 
      }
 }
   \vspace{-0.2cm}
   \caption{Challenge Categories for Image and Video}  
   \label{tab:challenge_categories}  
    \vspace{-5mm}
\end{table}

\begin{table*}[h]
\centering
\renewcommand{\arraystretch}{1.2} 
\small
\begin{tabular}{lp{5.8cm}p{5.8cm}}
\toprule
\textbf{\centering Categorie} & \textbf{\centering Description} & \textbf{\centering Example Prompt} \\
\hline
\multicolumn{3}{c}{\textbf{Content}} \\
\hline
Human Activity & Descriptions about daily human activities, sports, performing arts, and professional skills. & A family enjoying a picnic in a park, children playing soccer.  \\
\hline
Animal Activity & Descriptions about wild animals, domestic pets, and interactions between animals. & A group of dolphins jumping out of the water.  \\
\hline
Natural Scenes & Descriptions about weather changes, geological events, and astronomical phenomena. & A thunderstorm with lightning striking the ground.  \\
\hline
Artificial Scenes & Descriptions about cityscapes, interiors of buildings, vehicles, and industrial production. & A bustling city street with traffic and pedestrians.  \\
\hline
Physical Phenomena & Descriptions about physical occurrences like candle burning, ice melting, glass breaking, and explosions. & A glass shattering in slow motion.  \\
\hline
Story & Descriptions about coherent narratives based on a story or fantasy rather than a single scene or activity. & Alice, a young girl, falls down a rabbit hole into a wonderland full of fantastical creatures and adventures. \\
\hline
Other & Descriptions about various contents that do not fit into the other specified categories. & Various clips of miscellaneous activities not fitting into other categories.  \\
\hline
\multicolumn{3}{c}{\textbf{Challenge}} \\
\hline
Style & Descriptions about artistic styles such as realistic, cyberpunk, and animated. & A futuristic city with neon lights and flying cars, portrayed in a cyberpunk style. \\
\hline
Color/Tone & Descriptions about color schemes like warm tones, cool tones, monochrome, and high saturation. & A serene landscape in warm, golden tones during sunset.  \\
\hline
Camera Movement & Descriptions about different camera movements, including fixed, panning, zooming, tracking, and aerial shots. & A drone shot capturing a bird's eye view of a mountain range. \\
\hline
Special Effects & Descriptions about special effects such as particle effects, lighting effects, and transitions. & Fireworks exploding with sparkling particle effects.  \\
\hline
Material/Texture & Descriptions about materials and textures like metal, wood, glass, and fabric. & Close-up shot of rain droplets on a glass window.  \\
\hline
Surreal & Descriptions about dreamlike, fantastical, or non-realistic elements. & A dreamlike scene with floating islands in the sky.  \\
\hline
Temporal Speed & Descriptions about different speeds, including slow motion, normal speed, fast motion, and time reversal. & Slow-motion capture of a hummingbird in flight.  \\
\hline
Spatial Relationships & Descriptions about the spatial arrangement of objects, their sizes, occlusions, and perspectives. & A house of cards being built, showing each layer's spatial arrangement. \\
\hline
World Knowledge & Descriptions about physical laws, famous landmarks, historical events, and renowned personalities. & A documentary about the pyramids of Egypt.  \\
\hline
Logical Consistency & Descriptions about ensuring logical relationships among events, timelines, and spatial layouts. & A mystery story where clues are pieced together logically. \\
\hline
Emotional Expression & Descriptions about expressions of emotions such as joy, sorrow, fear, and surprise. & A close-up of a person expressing joy after receiving good news.  \\
\hline
Text & Descriptions about incorporating textual elements dynamically within the footage. & An animated title sequence with dynamic text effects.  \\
\bottomrule
\end{tabular}
\caption{Video classification standards with example prompts.}
\label{tab:monet_detail}
\end{table*}

\clearpage



\end{document}